\documentclass[10pt,journal]{IEEEtran}
\usepackage[utf8]{inputenc}
\usepackage{textcomp}
\usepackage[pdftex]{graphicx}
\usepackage{booktabs}

\DeclareGraphicsExtensions{.pdf,.jpeg,.png,.eps}
\usepackage{epstopdf}
\usepackage{microtype}
\usepackage{xurl} % or \usepackage{url}
\usepackage{float}

\usepackage{algpseudocode} 
\usepackage[frenchb, english]{babel}
\usepackage{csquotes}
\usepackage{cite}
\usepackage{caption}
\usepackage{subcaption}
\usepackage[cmex10]{amsmath}
\usepackage{float,multirow}
\usepackage{amsfonts}
\usepackage[none]{hyphenat}
\usepackage{url}

\usepackage{breakurl}
\usepackage[breaklinks]{hyperref}
\usepackage{soul}
\usepackage{dirtytalk}
\usepackage{hyperref}
\usepackage{comment}
\usepackage[euler]{textgreek}
\usepackage{epsfig,bm,amsmath,amssymb,graphicx,setspace}
\usepackage{algorithm}
\usepackage{color}
\usepackage[T1]{fontenc}
\usepackage{algpseudocode}
\usepackage{textgreek}
\usepackage{amsthm}
\usepackage{array}
\usepackage{balance}
\usepackage{enumerate}
\usepackage[utf8]{inputenc}
\usepackage{array}
\usepackage{wrapfig}
\usepackage{multirow}
\usepackage{tabularx}
\usepackage[section]{placeins}
\usepackage{float}
\usepackage{placeins}

\usepackage{subcaption}
\usepackage[utf8]{inputenc}
\usepackage[export]{adjustbox}
\usepackage{wrapfig}
\usepackage{array}
\usepackage{multirow}
\usepackage{graphicx}

\usepackage[table,xcdraw]{xcolor}

\def\BibTeX{{\rm B\kern-.05em{\sc i\kern-.025em b}\kern-.08em
        T\kern-.1667em\lower.7ex\hbox{E}\kern-.125emX}}
\usepackage[T1]{fontenc}
\usepackage{multirow}
\usepackage[frenchb]{babel}
\newcolumntype{C}[1]{>{\centering\arraybackslash}m{#1}}
\newcolumntype{P}[1]{>{\centering\arraybackslash}p{#1}}
\usepackage{graphicx}
\usepackage{graphicx}
\usepackage{tikz}
\usetikzlibrary{calc,positioning}
\usepackage{stfloats} % allows bottom placement for figure*
\usepackage[T1]{fontenc}

\definecolor{wheelblue}{RGB}{0,92,170}
\definecolor{wheellight}{RGB}{224,237,255}
\definecolor{wheelring}{RGB}{245,250,255}

\tikzset{
  segtitle/.style = {font=\bfseries\small, wheelblue, align=center},
  outtext/.style  = {font=\footnotesize, align=left, text width=0.26\textwidth}, % width of bullet boxes
}
\begin{document}
  \title{Quantum Machine Learning for Cybersecurity: A Taxonomy and Future Directions}

\author{ Siva Sai, Ishika Goyal, Shubham Sharma, Sri Harshita Manuri, \\ Vinay Chamola~\IEEEmembership{Senior Member,~IEEE}, Rajkumar Buyya~\IEEEmembership{Fellow,~IEEE}

\thanks{Siva Sai and Rajkumar Buyya are with the Quantum Cloud Computing and Distributed Systems (qCLOUDS) Laboratory, School of Computing and Information Systems, The University of Melbourne, Australia (e-mail: \{sivasaireddy.naga, rbuyya\}@unimelb.edu.au).}

\thanks{Ishika Goyal is with Department of Computer Science and Engineering, Birla Institute of Technology and Science, Pilani, Pilani Campus, Vidya Vihar, Pilani, Rajasthan 333031, India
(e-mail: f20220549@pilani.bits-pilani.ac.in).}

\thanks{Shubham Sharma and Vinay Chamola are with the Department of Electrical and Electronics Engineering, Birla Institute of Technology and Science, Pilani, Pilani Campus, Vidya Vihar, Pilani, Rajasthan 333031, India (e-mail: \{p20240036, vinay.chamola\}@pilani.bits-pilani.ac.in). Vinay Chamola is also with APPCAIR.
}
\thanks{Sri Harshita Manuri is with Department of Computer Science and Engineering, Birla Institute of Technology and Science, Pilani, Hyderabad campus, India
(e-mail: f20231374@hyderabad.bits-pilani.ac.in).}

\thanks{This work was supported by the CHANAKYA Fellowship Program of TIH
Foundation for IoT \& IoE (TIH-IoT) received by Dr. Vinay Chamola under
Project Grant File CFP/2022/027.}
}
\maketitle
\begin{abstract}
The increasing number of cyber threats and rapidly evolving tactics, as well as the high volume of data in recent years, have caused classical machine learning, rules, and signature-based defence strategies to fail, rendering them unable to keep up. An alternative, Quantum Machine Learning (QML), has recently emerged, making use of computations based on quantum mechanics. It offers better encoding and processing of high-dimensional structures for certain problems. This survey provides a comprehensive overview of QML techniques relevant to the domain of security, such as Quantum Neural Networks (QNNs), Quantum Support Vector Machines (QSVMs), Variational Quantum Circuits (VQCs), and Quantum Generative Adversarial Networks (QGANs), and discusses the contributions of this paper in relation to existing research in the field and how it improves over them. It also maps these methods across supervised, unsupervised, and generative learning paradigms, and to core cybersecurity tasks, including intrusion and anomaly detection, malware and botnet classification, and encrypted-traffic analytics. It also discusses their application in the domain of cloud computing security, where QML can enhance secure and scalable operations. Many limitations of QML in the domain of cybersecurity have also been discussed, along with the directions for addressing them.
\end{abstract}

\section{Introduction}

% The rapid expansion of digital infrastructure has improved connectivity, productivity, and service delivery, but it has also enlarged the attack surface and heightened exposure to sophisticated cyber threats. Contemporary economies depend on tightly coupled systems—critical infrastructure, global financial rails, healthcare platforms, cloud services, and billions of consumer and industrial IoT devices—whose interdependencies propagate failures and widen adversarial footholds. Attack operations now span large-scale data breaches and targeted campaigns, including state-sponsored advanced persistent threats (APTs), with increasing persistence, operational security, and automation. In parallel, commoditised offensive tooling and the broad availability of artificial intelligence (AI) have shortened attacker development cycles and accelerated the emergence of novel intrusion techniques. Rule-based controls, signature matching, and conventional machine-learning pipelines are often outpaced by adaptive adversaries that manipulate features, blend with background traffic, or shift tactics across kill-chain stages. This imbalance between offensive agility and defensive capacity calls for architectures that deliver low-latency detection, strong generalisation to previously unseen behaviour, and scalable operation under realistic compute and energy budgets\cite{r1,r2}.
Digital infrastructure has expanded rapidly in the last decade. While this has led to better connectivity and delivery of service, it has also resulted in increased exposure to complex cyber threats. In modern economies, various systems such as critical infrastructure, global financial rails, healthcare platforms, cloud services, and billions of consumer and industrial IoT devices are all tightly interdependent.  These interdependencies make them more prone to the propagation of failures and adversaries.  Large-scale data breaches and targeted operations are common in recent attack operations, including state-sponsored advanced persistent threats (APTs), with increasing persistence, operational security, and automation. Furthermore, the availability of offensive tools and artificial intelligence has shortened development cycles of attackers. Novel intrusion techniques take less time and effort to develop. Rule, signature, and classical machine learning based approaches cannot keep up with the adapting adversaries. These adversaries can manipulate features, blend with background traffic and shift their tactics across kill-chain stages. The defensive capacities are severely outmatched in front of the agility of the attackers. Thus, architectures that deliver low-latency detection, strong generalisation to previously unseen behaviour, and scalable operation under realistic compute and energy budgets are needed \cite{r1}\cite{r2}.

% QML integrates quantum computation with statistical learning to address high-dimensional pattern recognition and decision-making tasks. By encoding data into parameterised quantum circuits and exploiting superposition and entanglement, QML can realise expressive
% feature maps and fast kernel evaluations that are difficult to emulate classically under certain regimes \cite{r3}. In cybersecurity, where signals are noisy, nonstationary, and often adversarially obfuscated, such representations can improve discrimination between benign variability and subtle malicious behaviour. When coupled with classical preprocessing and post-processing, hybrid quantum–classical pipelines offer a path toward adaptable and scalable analytics. A recent market analysis estimates the quantum–AI malware-detection segment at USD 1.92 billion in 2024 and projects a 29.4\% compound
% annual growth rate (CAGR) to USD 16.6 billion by 2033.
% These projections reflect growing commercial interest and the perceived limits of purely signature- and rule-based defences, motivating a systematic assessment of where QML yields measurable gains over strong classical baselines \cite{r26}.

 QML combines statistical learning with quantum computation. It performs especially well for high-dimensional pattern recognition and decision-making tasks.  Some QML pipelines use a combination of quantum and classical methods for preprocessing and postprocessing steps. QML encodes data using parameterised quantum circuits, leveraging superposition and entanglement that result in rich feature maps and fast kernel evaluations, which are hard to emulate classically \cite{r3}. In the domain of cybersecurity, where signals are noisy, non-stationary, and maliciously distorted, QML representations can improve discrimination between benign variability and subtle malicious behaviour. 
 % The market of malware detection using quantum-AI techniques is estimated to be valued at USD 1.92 B (2024), and is projected to reach USD 16.6 B by 2033. 
 The limitations of traditional signature and rule-based defences are the reason for this interest in QML-based techniques\cite{r26}. Thus, we present a systematic evaluation of QML techniques against strong classical baselines in this work.

\begin{figure*}[!t]
\centering
\includegraphics[width=\textwidth]{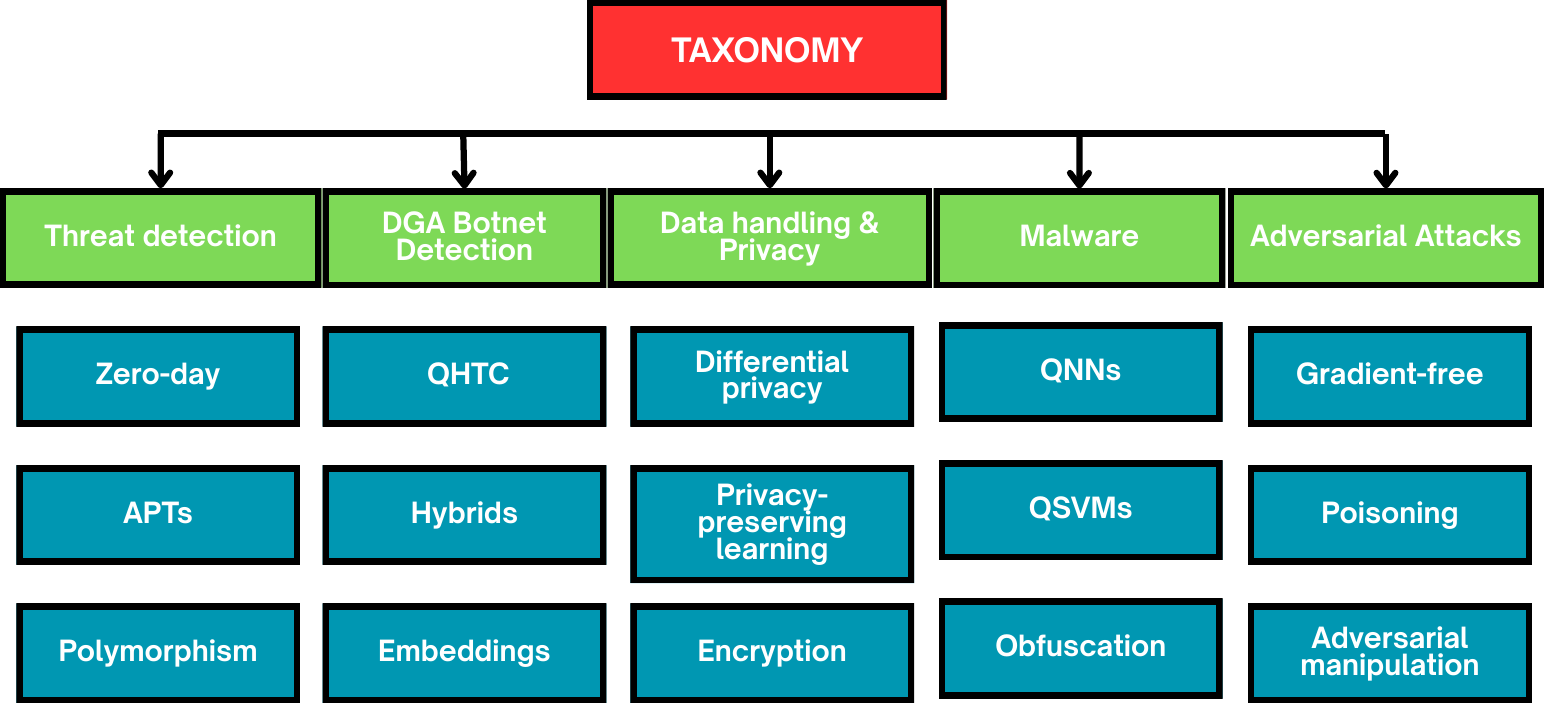}
\caption{Taxonomy of QML algorithms for cybersecurity}
\label{fig:tax}
\end{figure*}

% A principal motivation for QML is its potential to address high-dimensional inference and optimization with fewer effective computational steps than strong classical baselines under specific regimes \cite{r5}.  In cybersecurity telemetry, including network flows, system logs, endpoint events, and threat-intelligence feeds, the volume, velocity, and heterogeneity strain conventional pipelines. Hybrid quantum–classical models can leverage expressive feature maps and kernel evaluations to surface weak cross-modal correlations and rare-event structure in near real time. As adversaries adopt code obfuscation, multi-stage social engineering, and AI-enabled automation, detectors must generalise under distribution shift and adversarial manipulation. Early studies indicate that quantum-enhanced classifiers can flag anomalies in encrypted traffic using flow-level statistics without payload decryption, thereby preserving privacy while retaining detection utility \cite{r4}.  Quantum generative models further enable the synthesis of realistic attack traces and counterfactual scenarios for red-team training and pre-deployment stress testing. Beyond detection, QML offers formulations for combinatorial and stochastic optimisation—such as key-rotation scheduling, attack-graph path forecasting, and budget-constrained allocation of defensive resources across cloud, edge, and large-scale Internet of Things (IoT) tiers—supporting timely, resource-aware decisions at scale \cite{r8,r24}.
QML has promising potential in addressing high-dimensional inference and optimisation with fewer effective computational steps than strong classical baselines\cite{r5}. Conventional pipelines in cybersecurity telemetry struggle to keep pace with the volume, speed, and diversity of data generated from network flows, system logs, endpoint events, and threat intelligence feeds. A hybrid approach of quantum and classical techniques can leverage rich feature maps and kernel evaluations. This can reveal weak cross-modal correlations and rare-event structures in near real-time. Adversaries use techniques such as code obfuscation, multi-stage social engineering, and AI-enabled automation. The detectors for these adversaries should generalise under distribution shift and adversarial manipulation. Previous studies have indicated that quantum-based classifiers, which utilise flow-level statistics without payload decryption, can effectively detect anomalies in encrypted traffic \cite{r4}. Thus, they are able to preserve privacy without the detection utility having to suffer.  Generative models based on quantum computing can simulate realistic attack traces and counterfactual scenarios, making them excellent for red-team training and pre-deployment stress testing. QML approaches are not only good at detection but also at combinatorial and stochastic optimisation, such as key-rotation scheduling and attack-graph path forecasting. They also shine in the timely allocation of defence resources across cloud edge and large-scale IoT tiers,  especially when there is a limited budget, and allocation needs to be done in a resource-aware manner\cite{r8}\cite{r24}. The taxonomy of QML algorithms for the domain of cybersecuerity have been summarized in Figure \ref{fig:tax}.

% Collectively, these capabilities position QML as a substantive enabler of next-generation cyber defence. This review undertakes a critical assessment of QML’s effectiveness in addressing current gaps in detection, response, and resilience, with emphasis on supervised, unsupervised, and hybrid quantum–classical formulations. It synthesises evidence from emerging deployments, delineates representative system architectures and data-encoding strategies, and characterises practical constraints arising from hardware noise and scale, feature-mapping overhead, and optimisation pathologies. In parallel, it outlines evaluation practices suited to security settings—including leakage-aware dataset splits, resource-accounted metrics, and robustness analyses against adaptive adversaries—to support credible comparisons with strong classical baselines. The objective is to identify when and why QML provides measurable gains, and to chart a research and deployment agenda for scalable, privacy-preserving, and operationally realistic security solutions.
% Figure \ref{fig:tax} represents the taxonomy for application section of this paper.

The combined capabilities highlighted above show how QML plays a key role in future advancements in cybersecurity. This review aims to critically evaluate the real-world impacts of QML in closing gaps in cyber-threat detection, incident response, and system resilience, covering supervised, unsupervised, as well as hybrid quantum-classical modes. It summarises findings from previous real-world and experimental QML deployments. It also identifies and explains the typical system architectures and quantum data-encoding methods. This review also discusses practical limitations that arise from things such as noise and scaling issues in current quantum hardware, the overhead caused by feature-mapping processes, and optimisation difficulties specific to quantum models. It also discusses the evaluation framework and assessment practices specifically tailored for cybersecurity, such as leakage-aware dataset splits,
resource-accounted metrics, and robustness analyses against
adaptive adversaries that result in credible comparisons with strong classical baselines. The objective of this review is to determine the conditions in which QML can deliver substantial gains over baseline methods and to propose a research and deployment roadmap for scalable, privacy-aware, and practically deployable security solutions.

\section{Related Work}
% Quantum Machine Learning (QML) has recently emerged as a transformative paradigm that integrates the computational advantages of quantum computing with the adaptive learning capabilities of artificial intelligence to address complex cybersecurity challenges. Several recent studies have explored this intersection, offering insights into specific applications but differing in scope, methodological rigor, and depth of analysis.
QML has recently emerged as a novel and transformative technology, addressing numerous challenging problems associated with cybersecurity.  Integrating the computational advantages of quantum computing with the adaptive learning capabilities of artificial intelligence.  
QML spans across various domains. Recently, numerous studies have explored this technology, offering varying depths of insight.  

% Nguyen et al.~\cite{Nguyen2024} presented a broad survey of quantum computing applications in machine learning, emphasizing algorithmic innovations such as Quantum Support Vector Machines (QSVMs) and Quantum Neural Networks (QNNs). While their work provided a strong methodological foundation through a systematic review of 32 studies, it remained primarily theoretical and did not extend into domain-specific cybersecurity implementations.
In 2024, Nguyen et al.~\cite{Nguyen2024} presented a systematic review of 32 studies in QML. In their review, they did a thorough analysis of novel algorithms like QNNs and QSVMs for various applications. Their study, although methodologically strong, focused on the principles of the technology and did not specifically extend into the field of cybersecurity. 
% Ganapathy~\cite{Ganapathy2025} examined the role of QML in anomaly detection within cybersecurity audits, introducing algorithms such as QSVM, QNN, and Quantum Principal Component Analysis (QPCA). This work provided valuable insights into how quantum properties like superposition and entanglement can enhance pattern recognition; however, it was largely conceptual and lacked experimental benchmarking on large-scale or real-world datasets.
% Similarly, Dornala and Senthilkumar~\cite{Kiranmai2025} proposed an intelligent firewall-based malware detection framework that integrates Behavioral Analysis, QML, and Graph Neural Networks (GNNs). Their hybrid model achieved significant detection accuracy improvements, but the study primarily focused on architectural design rather than a systematic comparison of quantum algorithms or scalability across different cybersecurity domains.
Ganapathy~\cite{Ganapathy2025} focused their work in the domain of cybersecurity. They analyse algorithms such as QSVM, QNN, and Quantum Principal Component Analysis (QPCA), specifically for anomaly detection. They also analysed the properties of superposition and entanglement and their role in enhancing pattern recognition. Although their work explored the domain of cybersecurity specifically, it lacked experimental benchmarking on large-scale or real-world datasets. 
Dornala et al.~\cite{Kiranmai2025} proposed a framework, consisting of a combination of behavioural analysis along with Graph Neural Networks and QML, addressing intelligent firewall-based malware detection. The primary focus of their work was architectural improvements, and their framework achieved significant improvements in detection accuracy. However, it lacked a systematic comparison of algorithms based on quantum and classical methods, along with their degree of applicability on various fields of cybersecurity. 
% Eze et al.~\cite{Eze2025} focused on the comparative performance of classical and quantum models for malicious URL detection, evaluating QSVMs and Quantum Convolutional Neural Networks (QCNNs) against their classical counterparts. Although their results demonstrated the feasibility of quantum-enhanced models in improving classification accuracy, the study was limited to a single cybersecurity use case without discussing cross-domain generalizability or hybrid deployment considerations.
Eze et al.~\cite{Eze2025} explored the detection of malicious URLs. They aimed to analyze and compare the performance of classical and quantum-based approaches. In their study, they conclude that quantum approaches, namely QSVMs and Quantum Convolutional Neural Networks (QCNNs), outperform classical machine learning methods in classification accuracy. However, their study was limited to a single cybersecurity use case and could not explore the domain of cybersecurity as a whole, with its various subdomains and generalizability across them. Their study also lacked the consideration of systems that employ both classical and quantum approaches in a hybrid manner. Table \ref{tab:relatedqml} presents a comparison of this survey with the existing related surveys. 
% In contrast, the present review offers a unified and comprehensive synthesis of QML research in cybersecurity. It systematically integrates prior works across diverse threat landscapes, including intrusion detection, malware analysis, federated quantum learning, privacy-preserving encryption, and cloud security, while critically examining both algorithmic and infrastructural perspectives. Moreover, this paper consolidates existing QML paradigms (QSVM, QNN, QCNN, QGAN, and hybrid quantum–classical models) under a structured taxonomy, identifies practical gaps, and highlights emerging challenges in scalability, interpretability, and hardware dependence.

To address these shortcomings, this review presents a unified and comprehensive study of the application of QML in the domain of cybersecurity. This review aims to analyze previous works across diverse threat landscapes, including intrusion detection, malware analysis, federated quantum learning, privacy-preserving encryption, and cloud security, and critically examine both infrastructural and algorithmic perspectives. This paper presents a systematic consolidation of existing techniques, such as QSVM, QNN, QCNN, QGAN, and hybrid quantum–classical models, with a structured taxonomy while also identifying practical gaps, challenges in scalability, interpretability, and hardware dependence.

\begin{table*}[t]
\centering
\caption{Comparison of related surveys on Quantum Machine Learning (QML) in Cybersecurity}
\label{tab:relatedqml}
\renewcommand{\arraystretch}{1.15}
\small
\begin{tabular}{p{0.7cm} p{2cm} p{4cm} p{3cm} p{1.1cm} p{1.1cm} p{1.3cm} p{1.2cm}}
\toprule
\textbf{Ref.} & \textbf{Primary Focus} & \textbf{Key Contributions} & \textbf{Limitations} & 
\textbf{Hybrid} & \textbf{Tax.} & \textbf{Apps.} & \textbf{Bench.} \\
\midrule

\cite{Nguyen2024} & ML applications of Quantum Computing & 
Systematic survey of 32 studies integrating QC with ML; covers QSVM, QNN, QCNN, and hybrid learning frameworks & 
Broad and theoretical; limited domain-specific focus on cybersecurity or benchmark analysis &
\(\times\) & \(\circ\) & \(\circ\) & \(\times\) \\

\addlinespace[2pt]
\cite{Eze2025} & QML for Malicious URL Detection &
Comparative study of QSVM/QCNN vs.\ SVM/CNN on phishing and URL datasets; highlights early QML feasibility &
Single-task focus; lacks scalability, taxonomy, and broader cybersecurity coverage &
\(\times\) & \(\times\) & \(\circ\) & \(\checkmark\) \\

\addlinespace[2pt]
\cite{Ganapathy2025} & QML for Cybersecurity Audits &
Explores QNN, QSVM, and QPCA for anomaly detection; conceptual hybrid designs &
Theoretical orientation; lacks real-world validation and cross-domain integration &
\(\checkmark\) & \(\circ\) & \(\times\) & \(\times\) \\

\addlinespace[2pt]
\cite{Kiranmai2025} & Intelligent Firewall with QML &
Hybrid behavioral analysis + QML + GNNs; improved malware detection &
Application-specific; limited taxonomy and inconsistent evaluation &
\(\checkmark\) & \(\times\) & \(\checkmark\) & \(\checkmark\) \\

\addlinespace[2pt]
\textbf{Our paper} & Comprehensive Review of QML for Cybersecurity &
Taxonomy of QML (QSVM, QNN, QCNN, QGAN); hybrid models; privacy, interpretability, scalability; cross-domain synthesis &
Empirical deployments left for future work &
\(\checkmark\) & \(\checkmark\) & \(\checkmark\) & \(\checkmark\) \\

\bottomrule
\end{tabular}

\begin{flushleft}
\footnotesize
\textit{Note:} 
\(\checkmark\) = fully addressed; 
\(\circ\) = partially addressed; 
\(\times\) = not addressed. 
\textbf{Hybrid} = inclusion of hybrid Quantum–Classical models; 
\textbf{Tax.} = presence of structured taxonomy; 
\textbf{Apps.} = application diversity across cybersecurity domains; 
\textbf{Bench.} = benchmarking or experimental validation.
\end{flushleft}
\end{table*}

\section{An overview of Quantum ML and Cybersecurity}
\label{sec1}
\subsection{Quantum ML}

% Quantum Machine Learning (QML) couples quantum computation with statistical learning to address high-dimensional inference and optimization under strict time and resource budgets. Rather than replacing established pipelines, QML augments them by using quantum states to represent and transform data through parameterized unitaries that exploit superposition and entanglement. Under specific assumptions about data structure and access models, such representations can yield learning procedures that are difficult to emulate classically.
Quantum Machine Learning combines quantum computation with classic statistical learning. It can deliver high-dimensional inference and optimisation under strict time and resource budgets. It doesn't propose entirely novel pipelines. Rather, QML involves augmenting existing pipelines using quantum states to represent and transform data through parameterised unitaries that use superposition and entanglement. The representations generated by QML can lead to learning procedures difficult to emulate using traditional methods under specific assumptions about data structure and access models.  

% In practice, most systems adopt a hybrid quantum–classical design. Classical routines handle data preprocessing, batching, and optimizer updates, while a quantum processor implements a feature map $U_{\phi}(x)$ and a trainable circuit $U_{\theta}$ (e.g., a variational quantum circuit). Training proceeds by minimizing a task loss through gradient-based or gradient-free methods, using estimators such as parameter-shift rules. Alternative formulations include quantum kernel methods (e.g., QSVM with implicit kernels induced by $U_{\phi}$) and probabilistic models based on quantum generative circuits. Design choices—encoding strategy (basis, angle, amplitude), circuit depth, entangling topology, and readout observables—jointly determine expressivity, trainability, and robustness \cite{r21}.
In practical applications, QML is generally coupled with classical methods in a hybrid quantum–classical design. Steps like data preprocessing, batching, and optimiser updates are offloaded to classical methods, whereas the quantum processor implements a feature map $U_{\phi}(x)$ and a trainable circuit $U_{\theta}$ (e.g., a variational quantum circuit). Training involves minimizing a task loss through gradient-based or gradient-free methods, using estimators such as parameter-shift rules. Some alternate formulations also exist, such as quantum kernel methods (e.g., QSVM with implicit kernels induced by $U_{\phi}$) and probabilistic models based on quantum generative circuits. The robustness and trainability are determined by various design parameters, such as the encoding strategy (basis, angle, amplitude), the depth of the circuit, the entangling topology, and the readout observables \cite{r21}.

%1 The interaction between the two fields is bidirectional. Quantum resources can accelerate subroutines central to learning (feature embedding, kernel estimation, or sampling), while machine learning contributes calibration, control, and error-mitigation techniques that stabilize noisy intermediate-scale quantum (NISQ) hardware. This feedback loop enables iterative improvement: better hardware widens the feasible model class, and better models tolerate realistic noise and device constraints.
%2 The relationship between the two technologies, namely machine learning and quantum computing, is reciprocal in a feedback loop. Quantum hardware can accelerate subroutines central to learning (feature embedding, kernel estimation, or sampling). Concurrently, the noisy intermediate-scale quantum (NISQ) hardware is stablizied calibration, control, and error-mitigation techniques that machine learning offers. This inturn helps the machine learning model iteratively. In other words, better hardware widens the feasible model class, and improved models better tolerate noise and device constraints.

Quantum hardware mechanics and machine learning enhance each other in a feedback loop. In machine learning, processes like feature embedding, kernel estimation, or sampling get accelerated by quantum hardware. This enhanced state of machine learning offers better error-correction, control, and calibration. This, in turn, contributes to the stabilization of NISQ hardware. The stabilised NISQ hardware again helps to improve the machine learning processes. This enhancement happens iteratively. Thus, in other words, better hardware improves the model, and the improved models perform better in removing noise and hardware constraints.

%1 Applications span simulation, combinatorial optimization, and pattern recognition across science and engineering. Of particular relevance to this work, security analytics benefit from expressive feature maps and fast similarity evaluations for rare-event detection and distribution-shift resilience, while optimization primitives support tasks such as attack-graph analysis and resource allocation. Although QML remains at an early stage of maturity, converging progress in hardware scale, coherence, and compiler toolchains together with clearer benchmarking against strong classical baselines continues to expand the set of problems where quantum–classical pipelines are competitive \cite{r25}.
%2 QML has various applications such as simulation, combinatorial optimization, and pattern recognition, across various domains in science and engineering. This work focuses mainly on its use in the domain of security analytics.  This domain benefits from expressive feature maps and fast similarity evaluations. This is significant for rare-event detection and staying robust to changes in data patterns. Additionaly, optimization methods in QML can help with important security tasks such as attack-graph analysis and resource allocation. QML is at an early stage of maturity. Yet, converging progress in hardware scale, coherence, and compiler toolchains, as well as better benchmarking against strong classical baselines continue to widen the scope of problems where quantum–classical pipelines are competitive \cite{r25}.
QML can be used across various domains, including, but not limited to, artificial intelligence, healthcare, finance, chemistry, materials science, cybersecurity, and climate modelling. It can be used for its abilities in faster computation and data analysis. It can also recognize complex patterns especially well. QML also has a specific use for simulation and combinatorial optimization. In this work, the focus is on the use of QML in the domains of cybersecurity and security analytics. QML offers fast optimizations, and leads to rich feature maps that significantly benefit the domain of cybersecurity. Also, QML can perform similarity evaluations very fast, making it especially suited for detecting anomalies or rare events, and thus staying robust in looking for changes in data patterns. Although the current state of QML is immature, ongoing advancements in quantum hardware capacity, qubit stability, and software toolchains have led to QML being increasingly competitive in a wider scope of domains.

\subsection{Cybersecurity}\label{method}

% Cybersecurity safeguards the confidentiality, integrity, and availability of digital assets across interconnected infrastructures. At scale, defenders must process high volume, heterogeneous telemetry, including network flows, authentication logs, endpoint events, and application traces, while meeting latency, privacy, and regulatory constraints. The detection problem is adversarial and nonstationary. Tactics, techniques, and procedures evolve, lateral movement is stealthy, and dwell times can be prolonged. These conditions render purely reactive controls insufficient and motivate analytic methods that anticipate, localize, and contain attacks before mission impact \cite{r4}.
% Cybersecurity is essential in protecting the digital assets of interconnected infrastructures.  It helps safeguard the confidentiality, integrity, and availability of these digital assets. When the scale of the data and assets is large, defenders must process high volume, heterogeneous telemetry, including network flows, authentication logs, endpoint events, and application traces, while not compromising on latency, privacy and resource budgets. The detection problem is adversarial and nonstationary.  It faces many challenges from the evolving procedures, techniques and tactics from the adversaries, rendering purely reactive controls insufficient. This motivtes the development of techniques that anticipate, localize, and contain attacks before impact\cite{r4}.
Cybersecurity is a crucial domain for managing assets in digital infrastructure. In digital systems such as networks, cloud services, and IoT, assets are interconnected, and a compromise in one component can easily spread to the entire system. Cybersecurity ensures the confidentiality, integrity, and availability of these digital resources. In recent years, the scale of operations and data has expanded substantially. There is now a wide variety of sources of telemetry data, including network traffic, user authentication logs, and endpoint data. Processing such high amounts of diverse and real-time data efficiently becomes a challenging task. The data must be processed with low latency in a privacy-preserving manner, all within the limited computing resources available in real-world scenarios. defence against cyber attacks is not static. It is a battle against continuously evolving intelligent adversaries. Fixed defences become obsolete as adversaries evolve their attack tactics and techniques, making reactive or signature-based approaches inadequate in such a nonstationary threat landscape. There is a need for security techniques that are proactive and anticipate the evolving nature of attacks and their tactics. Approaches that detect, predict, and respond before the attack causes harm, and provide early threat localization, attack containment, and resilience, are a must. These challenges motivate the development of adaptive, learning-based, and automated defence systems\cite{r4}.

A robust cybersecurity system is evaluated not just by its accuracy, but also by its ability to detect and contain threats in a timely manner. Time-to-detect (TTD) and Time-to-contain (TTC) are key indicators of performance. The system should also be able to handle unseen attack types and the tampering of inputs done by adversarial manipulations. Such a system must also keep false positives extremely low to avoid alert fatigue, while preserving privacy. Thus, we examine systems combining the quantum and classical techniques for cybersecurity. In complex or noisy datasets, the signals are weak. 
\section{How Does QML Enhance Cybersecurity?}\label{sec2}

Quantum computing leverages parallelism, enabling QML to compute rich similarity measures and sample complex distributions with great efficiency. This happens by encoding telemetry or features into quantum states (via quantum feature maps) and exploiting superposition and entanglement. When the data is high-dimensional and heterogeneous, this use of parallelism can lead to shorter training cycles. These advantages mean the QML is significantly more robust against complex and evasive threats, such as zero-day attacks and APTs, which conceal their activity and continually evolve to evade detection. Due to the above-mentioned advantages such as parallelism, quantum models can process large volumes of telemetry (network flows, system logs, endpoint traces) at scale, thus offering real-time identification of subtle or emerging anomalies and novel attack signatures. For example, even when adversaries camouflage their behaviour, QML classifiers can differentiate normal network fluctuations from early-stage intrusions using flow-level features.

QML is not limited to detection. It can provide proactive and reactive defences that are adaptive and resilient. Using QML, defensive policies can be dynamically updated in real time, accounting for concept drift (changes in normal behaviour) and shifting adversarial tactics. Furthermore, quantum generative models can create high-fidelity attack simulations for red-team exercises and what-if analyses.  Being both adaptive and generative in its capabilities, QML can support closed-loop operations across cloud, edge, and on-premise deployments. These automated security deployments result in lower TTD and TTC and fewer false positives due to the rich feature representations of QML.  QML also contributes to the modernization of cryptography and security. Post-Quantum Cryptography (PQC) targets encryption and key management layers, but it can still benefit from QML. PQC protects data at rest and in transit from quantum-capable adversaries, whereas QML strengthens live, real-time network defences through detection and prioritization. Combined, they ensure both data preservation (via PQC) and operational resilience (via QML) under practical constraints such as latency, energy, or compute limits. Thus, QML's benefit extends beyond reactive defence, providing proactive, learning-driven security.  Security provided by QML can scale with increasing data volume and complexity of adversarial attacks. 
Quantum feature mapping can amplify and detect such weak signals. QML can compute similarity evaluations and distances quickly and efficiently. It can detect anomalies in encrypted traffic without decrypting them by using the pattern in metadata or ciphertext. It can also generate synthetic attack traces, making it useful for red-team training and simulation of hypothetical scenarios. 

QML helps in Security Operations Centre (SOC) workflows by improving detection accuracy. It also improves the system's immunity against zero-day attacks. Hybrid QML setups, which combine quantum and classical methods at various stages of the pipeline, can be used to optimise computation, energy, and privacy constraints, which is crucial for operational environments. 
% This paper evaluates the practicality of such systems in real-world applications against strong classical baselines. It focuses on deployment patterns, benchmarks, and evaluation frameworks for credible and scalable adaptation of QML in cybersecurity applications\cite{r6}.
 % Table \ref{tab:classical_vs_qml} provides a comparative overview of classical and quantum machine learning approaches, highlighting their respective strengths, limitations, and implications for cybersecurity applications.
  Table \ref{tab:classical_vs_qml} compares classical and quantum machine learning approaches in their respective strengths and limitations. It also describes what each of these implies for cybersecurity applications. 

\section{Applications}%3-3.5 pgs
% The rapid evolution of cyber threats, combined with the growing complexity of digital infrastructures, has pushed traditional machine learning and rule-based systems to their limits. Applications of QML span multiple domains of cybersecurity, including threat detection, healthcare data security, malware classification, botnet identification, telecommunication protection, and adversarial defence. Each domain highlights unique challenges where QML models such as QSVMs, QNNs, Variational Quantum Classifiers (VQCs), and Quantum Generative Adversarial Networks (QGANs) demonstrate advantages over their classical counterparts in terms of speed, efficiency, and robustness. The following subsections present a detailed survey of these applications, illustrating how QML contributes to building resilient, future-ready security frameworks.
The applications of QML in the domain of cybersecurity are multifarious, such as adversarial detection and defence, healthcare data security, malware and botnet identification and telecommunication protection. Every domain presents unique challenges. Due to the growing complexity of digital infrastructures along with the rapid evolution of cyber threats, traditional machine learning and rule-based systems have been pushed to their limits.  In each of these domains, QML models such as QSVMs, QNNs, VQCs, and QGANs, have shown to be much faster, efficient, and robust as compared to their classical counterparts. In the following subsections, we present a detailed survey of these applications and show how QML contributes to future-ready and resilient security frameworks. Figure \ref{fig:tax} represents the taxonomy of the applications of QML in cybersecurity.

\begin{table*}[!ht]
\centering
\caption{Comparison of Classical vs Quantum Machine Learning Systems in Cybersecurity}
\label{tab:classical_vs_qml}
\renewcommand{\arraystretch}{1.15}
\small
\begin{tabular}{p{3.2cm} p{6.2cm} p{6.2cm}}
\toprule
\textbf{Aspect} & \textbf{Classical ML systems} & \textbf{Quantum ML systems} \\
\midrule

\textbf{Threat detection approach} &
Mostly reactive, identifies and responds after anomalies occur \cite{r4}. &
Proactive: predicts, identifies, and neutralises threats before impact \cite{r6}. \\

\addlinespace[2pt]
\textbf{Cryptographic capabilities} &
Uses standard encryption, vulnerable to quantum attacks. &
Supports quantum-safe encryption and adaptive quantum-secure models \cite{r25}. \\

\addlinespace[2pt]
\textbf{Computational power} &
Limited by classical hardware, struggles with high-dimensional problems. &
Uses quantum parallelism for large-scale optimisation and complex datasets \cite{r21}. \\

\addlinespace[2pt]
\textbf{Pattern recognition} &
Effective for known threats, may miss subtle anomalies. &
Detects hidden correlations, identifies zero-day and APT-level threats \cite{r7, r13}. \\

\addlinespace[2pt]
\textbf{Scalability} &
Performance declines with increasing data complexity. &
Quantum-enhanced scaling supports large, dynamic digital ecosystems \cite{r25}. \\

\addlinespace[2pt]
\textbf{Application domains} &
Used for IDS, malware detection, and threat monitoring \cite{r4}. &
Extends to cybersecurity, finance, healthcare, and climate modelling \cite{r21, r25}. \\

\bottomrule
\end{tabular}
\end{table*}

\subsection{Threat detection}
% As cyber threats grow in complexity, classical and conventional detection systems are increasingly struggling to keep pace. Modern attacks such as zero-day exploits, polymorphic malware, and advanced persistent threats (APTs) are deliberately engineered to bypass rule-based defences and, in many cases, to mislead even classical machine learning models. Classical rule-based and machine-learning frameworks are constrained by computational overhead and often lack the flexibility needed to track rapidly changing attack vectors. Quantum machine learning (QML) addresses these limitations by leveraging quantum resources to accelerate learning under specific regimes, support rapid model updates, and enable real-time anomaly detection and policy-driven response. In security contexts, these properties enable more durable defences that are better aligned with future threat dynamics.
Classical and conventional detection systems fail to keep pace with the exceedingly complex cyber threats. Rule-based defences, and in many cases, even classical machine learning models, are misled by modern-day attacks such as zero-day exploits, polymorphic malware, and advanced persistent threats, which are deliberately engineered to bypass these approaches. These approaches are also limited by computational overhead, where they lack the required acceleration to process the scale of computations. They also lack the flexibility needed to track evolving attack vectors.  QML is heavily accelerated due to its use of quantum computations, which also enables rapid model updates, allowing for real-time anomaly detection and policy-driven responses. All these properties make QML-based approaches much more durable and better aligned with future threat dynamics.

% The study in \cite{r2} examined QML-driven intrusion detection and reported high detection rates on the evaluated workloads: 99.7\% for known attacks and 96.4\% for zero-day attacks, with a false-positive rate of 0.03\%. The implementation sustained throughput up to 100,000 events per second and achieved a latency of 0.3~s, indicating potential for real-time detection and response at enterprise scale.
% To test the potential of QML-driven intrusion detection at a large scale, and both known and zero-day attacks, Awasthi\cite{r2} tested and reported high detection rates on evaluated workloads. They found a 96.4\% detection rate for zero-day attacks and 99.7\% for the known ones. The false positive rate also was just 0.03\%. These results indicate potential for real-time detection and response at enterprise scale.
To test the potential of QML-driven intrusion detection at a large scale, and both known and zero-day attacks, Awasthi \cite{r2} tested and reported high detection rates on evaluated workloads. They leveraged quantum parallelism to evaluate large sets of attack patterns simultaneously. The system used quantum-enhanced feature analysis and processed incoming network events. Thus, the system was capable of examining up to $2^{20}$ potential attack vectors in parallel. They benchmarked it against classical IDS systems for comparison. Their methodology involved measuring detection accuracy, latency, and false-positive rates across known and zero-day attack scenarios under identical conditions. Thus, they were able to perform a direct performance assessment between quantum and classical approaches. They found a 96.4\% detection rate for zero-day attacks and 99.7\% for the known ones. The false positive rate also was just 0.03\%. These results indicate potential for real-time detection and response at enterprise scale.
% Paper \cite{l1} proposed a hybrid quantum–classical detection framework targeting high-security deployments. Classical preprocessing mapped inputs to quantum states, enabling high-dimensional feature representations in which weak threat patterns became more separable. Quantum-supervised classifiers, including QSVMs and variational quantum circuits (VQCs), were then applied for classification and clustering by estimated threat likelihood. In simulation case studies, QSVMs surpassed classical SVMs on obfuscated threats, and VQCs exhibited greater sensitivity to faint behavioral anomalies than standard neural networks. These findings suggest that QML models can provide accuracy gains and enhanced anomaly sensitivity in high-risk cybersecurity settings.
Karamchand et al.\cite{l1} proposed a mixed approach of quantum and classical detection in their framework targeting high-security deployments. In their approach, classical pre-processing was used to transform inputs into a high-dimensional feature space in which weak threat patterns become more distinguishable. The transformed inputs were then fed into quantum-based classifiers such as QSVMs and VQCs for classification and clustering. Their evaluation on concealed threats showed that QSVMs were able to outperform their classical counterpart. Also, VQCs showed much better sensitivity to faint behavioural anomalies than what standard neural networks were able to achieve. These findings demonstrate the accuracy gains and enhanced anomaly awareness that QML models can provide in high-risk cybersecurity environments.

\subsection{Data handling and Privacy}
% Secure data management is a cornerstone of modern cybersecurity, particularly in sectors where sensitive information is constantly exchanged and stored. Quantum Machine Learning (QML) offers advanced techniques to protect healthcare records, encrypted communication streams, and other high-value datasets against evolving threats. By combining quantum-enhanced encryption, anomaly detection, and privacy-preserving learning models, QML enables more resilient and trustworthy data handling frameworks.
% A very fundamental aspect of model cybersecurity is secure data management. This becomes especially important in the fields where sensitive data is frequently exchanged or stored. To enhance the protection of this sensitive information, Quantum Machine Learning(QML) introduces various advanced techniques. QML can safeguard health records, encrypted communication channels, and other valuable datasets from the constantly evolving cyber threats. QML is especially good because it leverages quantum-enhanced encryption for stronger data protection and privacy-preserving learning models that allow model training without exposing private data. QML also integrates anomaly detection to identify unusual or malicious patterns in data activity. Due to all these factors, QML-based methods are more resilient, trustworthy, and secure frameworks for handling and managing data.
In modern cybersecurity, the secure management of data is a core component, particularly in domains that handle high-value and sensitive data, such as healthcare and communication systems.  QML can secure health records, encrypted communication channels, and other confidential datasets against evolving cyber threats by using quantum-enhanced encryption methods. QML can also utilise privacy-preserving learning methods, such as federated or encrypted learning, ensuring that the training process can occur without exposing the raw data. It can make data handling systems trustworthy and reliable by using these quantum-enhanced encryption techniques and privacy-preserving learning models.

Gupta et al.\cite{r4} proposed IQ-HDM, an intelligent quantum cybersecurity framework for healthcare data management. The aim of their proposed framework was to create an environment for managing healthcare data that is both secure and efficient. The proposed IQ-HDM framework comprises two modules: Quantum One-Time Padding Encryption (QOTPE) and Quantum Feed-Forward Neural Network (QFNN). QOTPE handled secure data storage by encrypting healthcare data into maximally mixed quantum states, while QFNN handled the detection of malicious intent and abnormal behaviour. This ensured that the data remains confidential while providing robust analysis of user behaviour patterns, allowing the framework to identify threats proactively and prevent incidents before damage occurs. The proposed framework achieved 89.02\% accuracy, significantly outperforming classical baselines by up to 67.6\% in performance.

% In \cite{l2}, a complementary healthcare-focused cybersecurity framework was presented that combines quantum machine learning with human factors. The design applies a local differential privacy layer followed by a Quantum Random Forest (QRF) to secure sensitive patient data while supporting learning on large, high-dimensional datasets. Experiments on the WUSTL\mbox{-}EHMS\mbox{-}2020 and ICU datasets confirmed effectiveness, achieving attack-detection accuracies of 99.90\% and 98.57\%, respectively, even in the presence of the privacy-preserving layer. These results indicate the potential of QML to deliver high accuracy together with robust privacy guarantees in electronic healthcare settings.
% In thier work, \cite{l2} proposed a healthcare-oriented cybersecurity framework integrating QML with human factors. They applied Local differential privacy(LDP) layer to protect sensitive patient data before the analysis begins. And after this privacy layer, a quantum random forest(QRF) model is used for secure learning from two large and high dimensional datasets. They tested their approach on two benchmark healthcare datasets, namely ICU dataset and WUSTL EHMS 2020 dataset. The outcome of their experiments was that they achieved 99.9\% attack detection accuracy on WUSTL EHMS 2020 dataset as well as 98.57\% attack detection accuracy on ICU dataset. The addition of the privacy preserving layer did not hinder the performance. These findings highlight that QML can achieve high detection accuracy even while maintaining strong privacy guarantees in electronic healthcare systems.
A recent study\cite{l2} focused on cybersecurity for healthcare systems, where sensitive patient data demands strict protection. This study examined how QML can be integrated with human-factor awareness to develop analytics pipelines that are more secure and trustworthy. In their framework, a Local Differential Privacy (LDP) mechanism is applied before any training occurs. This layer adds controlled noise or perturbation to the raw data, ensuring that individual patient records remain private even if compromised. This privacy preprocessing happens before the quantum learning stage begins, forming the first shield of defence. Then, the Quantum Random Forest (QRF) algorithm is used for secure pattern learning, enabling robust classification on large, high-dimensional datasets. Their approach proved to offer both speed and accuracy improvements compared to classical models, as the framework reached 99.9 \% detection accuracy on the WUSTL EHMS 2020 dataset and 98.57\% on the ICU dataset. The privacy layer did not degrade performance, proving that strong privacy guarantees and high detection accuracy can coexist. Their study demonstrates QML’s ability to maintain high detection accuracy while preserving data confidentiality in electronic healthcare systems.

\subsection{DGA Botnet Detection}

 % Domain Generation Algorithm (DGA) botnets generate large numbers of pseudo-random domain names for communication with command-and-control servers, which poses a challenge for classical detection and blacklist-based methods. QML provides alternative mechanisms to detect these patterns with improved timeliness and reliability.
 Domain Generation Algorithm (DGA) botnets create a large number of pseudo-random domain names to communicate with their command-and-control servers. Classical detection systems and blacklist-based methods struggle to effectively detect or block such domains. QML offers alternative mechanisms, which are able to recognise these dynamic and complex domain generation patterns, and thus QML-based detection improves both the speed and reliability of identifying DGA-driven botnet activities.

% In \cite{r7}, hybrid quantum-classical models were evaluated under NISQ constraints using simulators and devices from IonQ, Rigetti, Quantinuum, and the IBM AER simulator. On a DGA botnet dataset, the study trained Quantum Neural Networks (QNNs), Quantum Support Vector Classifiers (QSVCs), Pegasos, and Variational Quantum Circuits (VQCs), and introduced a Quantum Hoeffding Tree Classifier (QHTC). QHTC achieved the highest speed and accuracy and avoided optimization issues commonly observed in variational methods. The results indicate that, even under NISQ conditions, approaches such as QHTC can provide measurable advantages in cybersecurity applications.
% \cite{r7} assessed hybrid quantum–classical models under noisy intermediate-scale quantum (NISQ) hardware constraints. In their evaluation, they utilized both quantum simulators and actual devices from IonQ, Rigetti, Continuum, and the IBM AER simulator. They trained and compared multiple models on a DGA Botnet dataset. The models included Quantum Neural Networks (QNNs), Quantum Support Vector Classifiers (QSVCs), Pegasus Classifiers, and Variational Quantum Circuits (VQCs). They also introduced a new model, the Quantum Hoeffding Tree Classifier (QHTC). They found that QHTC achieved the highest accuracy and fastest performance among all models, while avoiding the optimization challenges faced by variational quantum models. Overall, the results demonstrate that even within NISQ limitations, models like QHTC offer practical and measurable benefits for cybersecurity tasks such as botnet detection.
In NISQ hardware, quantum noise and limited qubit counts constrain performance. Tehrani\cite{r7} examined hybrid learning approaches that combine quantum and classical methods to evaluate their effectiveness under these constraints for cybersecurity applications. The researchers used a combination of real quantum processors and simulators, including those from IonQ, Rigetti, and Continuum, along with the IBM AER simulator, and carried out experiments on the DGA Botnet dataset to evaluate intrusion and botnet detection. They compared several quantum and hybrid algorithms, such as QNNs, QSVCs, Pegasus classifiers, and VQCs, and also proposed a novel algorithm called the Quantum Hoeffding Tree Classifier (QHTC). QHTC outperformed all other models in terms of accuracy and speed while avoiding the optimisation issues common in variational circuit models. These results indicate significant gains from quantum-enhanced models, even when the hardware remains noisy and of small scale. Their study confirmed that, in real-world NISQ environments, models like QHTC are practical and secure for botnet and intrusion detection tasks.

% In \cite{r8}, a hybrid quantum-classical deep learning model combined PennyLane Angle and IQP embeddings with circuit layouts including Basic Entangler, Random, and Strongly Entangling Layers. Applied to selected DGA botnet features, the model achieved up to 94.7\% accuracy, indicating the utility of quantum-enhanced methods for specialized threat detection tasks.
% In \cite{r8}, the authors introduced a hybrid quantum–classical deep learning model specifically designed for detecting domain-generating algorithm (DGA) botnet activity. They integrated PennyLane angle embeddings and instantaneous quantum polynomial (IQP) embeddings for feature encoding. Different quantum circuit layouts were explored, including basic entangler layers, random layers, and strongly entangling layers, which were applied to selected DJI botnet features for classification. The proposed approach achieved an accuracy of up to 94.7\%, demonstrating the effectiveness of quantum-enhanced techniques for such detection tasks.
Suryotrisongko et al. \cite{r8} developed a hybrid deep learning framework combining quantum and classical components to detect botnet activity generated by domain-generating algorithms. Botnets created through such algorithms are particularly difficult to identify. The framework incorporated quantum feature encoding techniques using PennyLane angle embeddings and instantaneous quantum polynomial (IQP) embeddings, which enabled the mapping of classical features into quantum state representations. The authors experimented with multiple circuit architectures, including basic entangler layers, random quantum layers, and strongly entangling layers. The hybrid model achieved up to 94.7\% accuracy, significantly outperforming classical baselines. They concluded that quantum-enhanced embeddings and circuit designs are especially effective for botnet detection and broader cybersecurity analytics.
% Tehrani et al. \cite{r9} proposed a hybrid quantum machine learning method for DGA botnet detection, the Quantum-enhanced Hoeffding Tree Classifier (QHTC). The pipeline processed online data streams using Google Cloud and Azure with Azure Quantum providers (Rigetti, IonQ, Quantinuum), together with the Qiskit library and the Aer simulator. Reported results included 100\% accuracy in the final test round and a mean accuracy of 91.2\% on 5{,}000 samples, with a total execution time of 1{,}687 seconds, improving on prior simulator-only results of 76.8\% accuracy on 1{,}000 samples.
% \cite{r9} introduced a hybrid quantum machine learning approach for DGA botnet detection. They use the Quantum Enhanced Hoeffding Tree Classifier (QHTC), designed to handle online data streams efficiently. The implementation was deployed using Google Cloud and Microsoft Azure infrastructures, integrating with Azure Quantum providers such as Rigetti, IonQ, and Quantinuum. The framework utilized Qiskit along with the IBM AER simulator. In their experiments, the authors achieved 100\% accuracy in the final testing round and recorded a mean accuracy of 91.2\% on 5,000 samples, with a total execution time of 1,687 seconds. This performance significantly outperformed earlier simulated-only experiments, which achieved only 76.8\% accuracy on 1,000 samples. 
Tehrani et al. \cite{r9} used a QHTC for detecting domain generation algorithm botnets. They employed a hybrid approach that utilised both quantum and classical methods to enhance the accuracy and efficiency of real-time detection in network environments. They used public cloud infrastructure such as Google Cloud and Microsoft Azure, along with quantum providers like Rigetti, IonQ, and Quantinuum. The IBM Qiskit framework and AER simulator were used for quantum circuit design and experimentation. They found that the model achieved a perfect accuracy of 100\% in the final testing run. Across 5,000 test samples, it reached an average accuracy of 91.2\%, completing the entire process in just 1,687 seconds. Previous experiments had been limited to simulated runs only, where accuracy plateaued at around 76.8\% on a smaller dataset of 1,000 samples. Thus, a substantial gain in performance and scalability is indicated when moving from simulation-only setups to real hybrid cloud deployments. These results validate the feasibility of integrating quantum-enhanced classifiers with cloud-based cybersecurity systems.

\begin{figure*}[h]
    \centering
    \includegraphics[width=\textwidth]{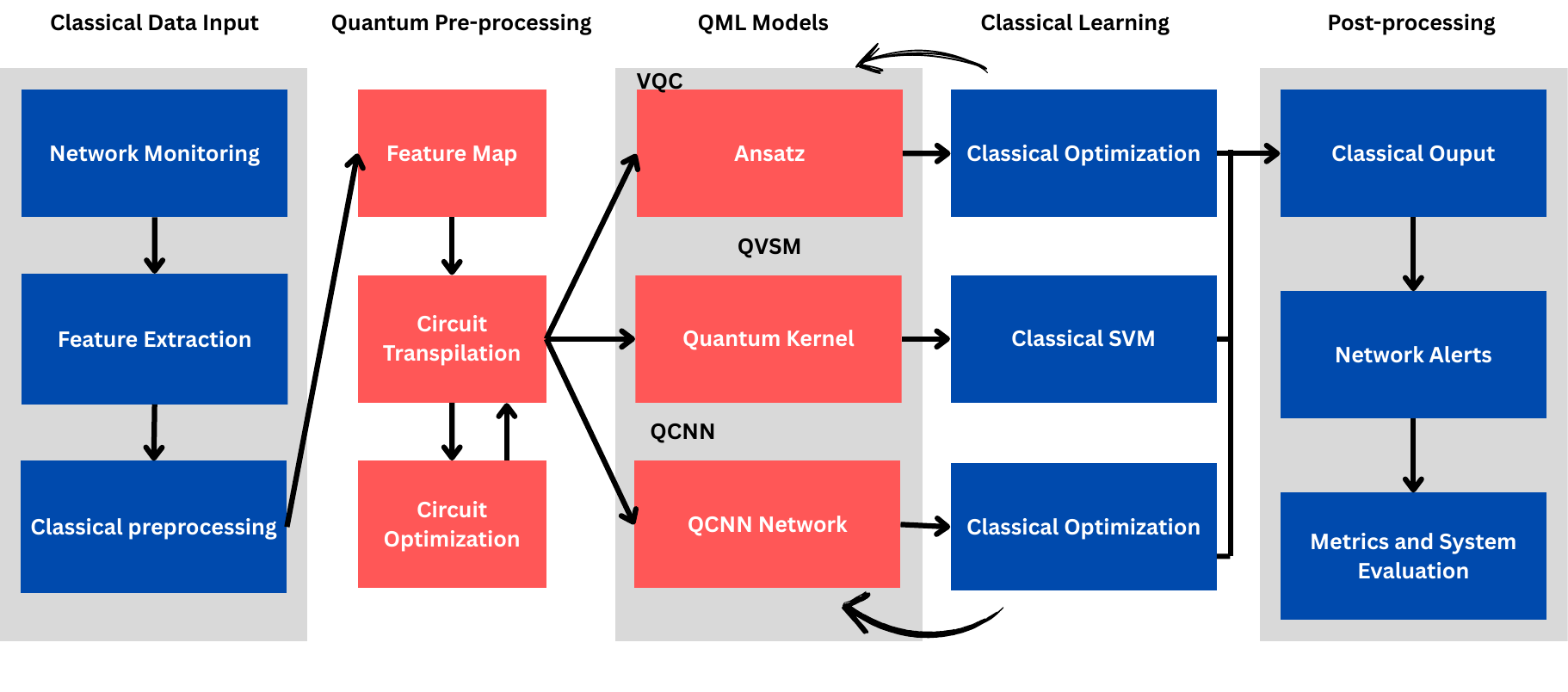}
    \caption{Operational flowchart of QML-IDS showing end-to-end integration of classical and quantum components, including network monitoring, quantum circuit optimization, QML model execution (VQC, QSVM, QCNN), and classical post-processing for threat detection and evaluation.}
    \label{fig:QMLIDS_flowchart}
\end{figure*}
Abreu et al.\cite{r18} introduced QML-IDS, combining classical methods with multiple quantum models such as VQCs, QSVMs, and QCNNs for enhanced intrusion detection. They evaluated the framework on IBM Kyoto and IBM Osaka NISQ hardware backends. They found that QML-IDS was able to outperform classical baselines in both binary and multiclass classification. The QCNN model achieved a high F1-score of 95.60\% on the CIC-IDS-17 dataset for binary classification and 98.88\% for multiclass analysis attack classification on the UNSW-NB-15 dataset. Although they were not able to significantly outperform classical methods in all aspects, they were equal to or superior to traditional methods. Figure \ref{fig:QMLIDS_flowchart} shows the working of the QML-IDS system.
\subsection{Malware}
 QML has been recently explored in the domain of malware detection and containment. Malware evolves rapidly, and QML can offer adaptive detection, improving detection accuracy that traditional static systems are unable to achieve. Bikku et al.\cite{r10} proposed a QNN-based framework for detecting malware was proposed. It consisted of three phases: feature extraction, classification, and real-time analysis, all of which were achieved through quantum methods. Quantum-based feature extraction was used to capture malware characteristics, and then a QNN was used for differentiating between malicious and benign samples. This framework was tested on the MalwareDB dataset, and it achieved 95\% accuracy, not only outperforming the classical machine learning models such as Support Vector Machines (SVMs) and Random Forests, but quantum baselines such as quantum SVMs and quantum decision trees, all while having a minimal latency.

 Barrué et al.\cite{r11} analysed several classical algorithms with their quantum counterparts, for generalizable and robust malware detection. They compared QSVMs and QNNs with classical SVMs and neural networks. They tested two different preprocessing approaches. In the first approach, they performed executable-to-image transformation, converting binary executables into image representations. In the second approach, they used Ordered Importance Features (OIF) extraction, focusing on statistics across different attributes. They achieved 80.75\% accuracy using a QSVM with the $ZZphiFeatureMap$ on the image-based dataset, outperforming classical SVMs, which achieved 77\% accuracy with linear and polynomial kernels. On the OIF dataset, the QSVM with the $ZZpFeatureMap$ reached 95\% accuracy, comparable to the best classical SVMs. QNNs could not outperform classical models in accuracy, but optimization methods such as Simultaneous Perturbation Stochastic Bifurcation and data re-uploading improved runtime efficiency on the same accuracy. 

% In \cite{r12}, Quantum Support Vector Machines (QSVMs) were evaluated against classical SVMs on raw-binary datasets using Qiskit simulators and IBM quantum hardware. The results highlighted both potential and current challenges, including circuit transpilation overhead and job-size limits, and emphasized the need for custom adjustments to toolchains such as Qiskit. The study indicated that QSVM approaches are increasingly viable for real-world malware-detection applications as tooling and hardware mature.
% In \cite{r12}, quantum support vector machines (QSVMs) were compared with classical SVMs on raw binary malware datasets, conducted using Qiskit simulators and IBM quantum hardware. The results showed promising performance, demonstrating the potential of QSVMs, but also highlighted some challenges. Circuit transpilation overhead increased computation time. Job size limitations on quantum hardware meant that these experiments could not be performed on a large scale. Also, the need for customized adjustments was there to existing toolchains such as Qiskit to improve compatibility and efficiency. The study concluded that with the evolving hardware and software quantum infrastructure , QSVM-based approaches are becoming more and more practical and applicable for real-world malware detection tasks.
 In Kr{\'a}tk{\'a} et al.\cite{r12} the performance of classical SVMs were compared with QSVMs for binary malware detection. The experiments showed the performance boost achieved in the quantum-based approach over the classical approach, but also highlighted some challenges experienced in the former. Compuation time was increased due to Circuit transpilatiom and job size limitations on quantum hardware meant that these experiments could not be performed on a large scale. Also, the current evaluation infrastructure lacked the efficiency and compatibility to run these experiments on scale. The study concluded that QSVM-based approaches are bound to become increasingly practical and applicable for real-world malware detection tasks as  hardware and software quantum infrastructure evolves. 

\subsection{Anomaly Detection}
% Anomaly detection plays a critical role in cybersecurity by identifying unusual patterns that may indicate malicious activity. As QML approaches have come up, researchers are exploring new ways to enhance detection accuracy and efficiency beyond classical approaches. This subsection reviews recent advancements in QML-based anomaly detection methods and their applications.
% A core component of cybersecurity is anomaly detection, which aims at identifying unusual or suspicious patterns that may indicate malicious activity. QML inspires new approaches to enhance both accuracy and efficiency beyond what classical detection techniques can achieve in anomaly detection. QML’s ability to model non-linear and high-dimensional relationships gives it an edge in complex tasks. In this subsection, a review is provided for recent advancements in QML-based anomaly detection methods. This subsection also highlights their practical applications in cybersecurity.
In the context of cybersecurity, an anomaly refers to unusual or suspicious patterns that may indicate malicious activity. Anomaly detection involves the use of detection techniques to accurately identify such patterns. QML has resulted in novel approaches that enhance both accuracy and efficiency beyond what classical detection techniques can achieve. This advantage stems from QML's ability to model non-linear and high-dimensional patterns.  In this subsection, a review is provided for recent advancements in QML-based anomaly detection methods. This subsection also highlights their practical applications in cybersecurity.

With their advent, QML models have shown impressive performance in terms of accuracy and latency. Hossain et al.\cite{r1} integrated QNNs and QSVMs in a hybrid pipeline for real-time monitoring of network traffic. The QNN achieved 96\% accuracy in anomaly detection, and the QSVM achieved 93\% accuracy, with the average alerting latency being near-instant at 10 ms. Exploring further, Noah et al.\cite{r14} ventured in QML for cybersecurity in 5G networks. They used hybrid simulations using the IBM Qiskit and TensorFlow Quantum frameworks. Their quantum-assisted intrusion detection system, based on QNNs, achieved 96.8\% accuracy, whereas the classical approaches scored only  achieved only 92.3\%. Their QSVM was able to achieve an F1 score of 0.94 for zero-day attack detection, whereas classical models achieved just 0.87\%. Also, the detection latency got reduced from 320 ms to 110 ms with quantum processing, which is a critical improvement in this domain.  Still, the authors noted that hardware noise and limitations, interoperability issues, and cost remain hindrances to the adoption of this technology. 

\begin{figure*}[!t]
    \centering
    \includegraphics[width=0.5\textwidth]{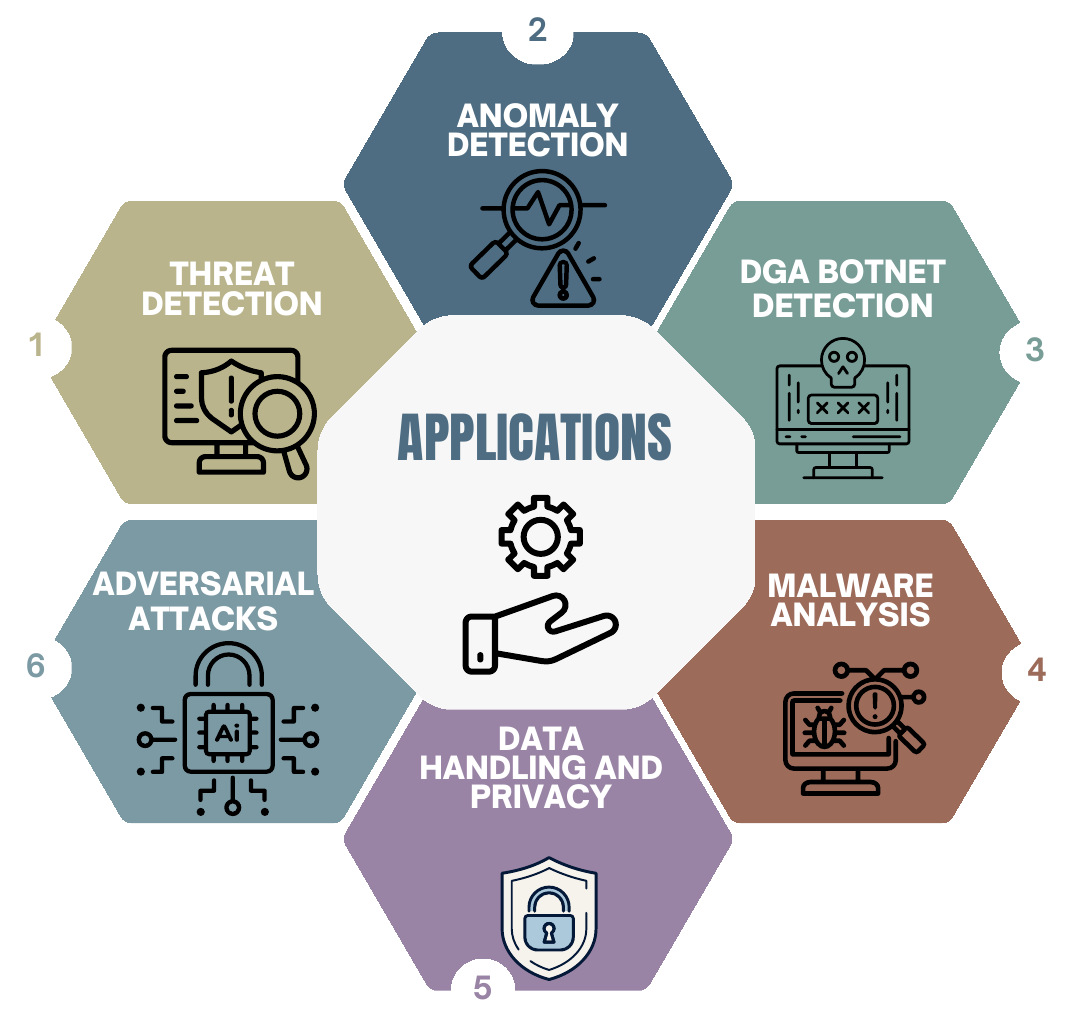}
    \caption{Application landscape of Quantum Machine Learning in cybersecurity.}
    \label{fig:qml_applications}
\end{figure*}

% In \cite{r13}, the authors argued that QML algorithms can exploit quantum parallelism and entanglement to process large datasets more efficiently than classical models, enabling faster and more accurate anomaly detection. The paper discussed cybersecurity applications, including quantum-enhanced anomaly detection using quantum kernels and QSVMs for analyzing encrypted traffic, and highlighted ongoing work with QNNs and hybrid quantum-classical architectures that combine classical deep learning with quantum computation for adaptive, real-time security. The authors concluded that as quantum hardware matures, QML will play a central role in securing telecommunication infrastructures against classical and quantum-enabled threats.
Quantum hardware is not used only in defence, but also as an offensive weapon. Thus, QML is instrumental in not just safeguarding against classical threats, but quantum-enhanced ones. In their work, Oyebode et al.\cite{r13} report that as quantum hardware advances, QML is expected to become central to securing telecommunication infrastructures against both classical and quantum-enabled threats. QML can use quantum parallelism and entanglement to process large datasets more efficiently. The paper discussed cybersecurity applications, including quantum-enhanced anomaly detection using quantum kernels and QSVMs for analysing encrypted traffic. They also discussed emerging approaches that integrate QNNs with classical deep learning in hybrid architectures.

\subsection{Adversarial Attacks}
Robust detection frameworks are crucial against adversarial attacks that are gradient-free. In their work, Kejriwal et al.\cite{v3} proposed a quantum-based defence framework for defence against such attacks,  combining classical components with QSVMs based on quantum kernels. They evaluated the framework on a combination of malware detection and network intrusion datasets. In their analysis, classical deep models suffered more than 85\% attack success, while the quantum-based methods reduced the vulnerability by up to 60\%. Similarly,  Ergu et al. \cite{h3} explored 6G interference classification by integrating QML with Open Radio Access Network architectures. They proposed a Hybrid Quantum–Classical Neural Network (HQ-CNN).  They used  ResNet-18 for deep feature extraction and Principal Component Analysis for dimensionality reduction. By using Q-Poison attacks, they manipulate network state classifications. Their HQCNN was able to achieve an accuracy of 78.5\% on clean data, but sees a 1.44x drop to just 54.6\% when fed with the attacks. Thus, defences against adversarial threats that exploit quantum coherence need to be strengthened further. 

Figure~\ref{fig:qml_applications} illustrates the range of QML applications in cybersecurity. 
% Table~\ref{tab:QML_summary} provides a summary of the domains, techniques, and performance improvements reported across these use cases.

% \section{ML techniques that can be used  in QML for Cybersecurity applications}
\section{ML techniques-based classification}
% Machine Learning serves as the foundation for both classical and quantum approaches in cybersecurity. ML learns from data, detect hidden patterns, and adapt to evolving threats. Within QML, these methods are done through quantum mechanics, allowing models to operate in higher-dimensional feature spaces, optimize more efficiently, and, in some settings, achieve greater accuracy in classification or clustering tasks. Core ML paradigms, including supervised learning, unsupervised learning, and generative modeling, extend into the quantum domain, resulting in algorithms such as Quantum Support Vector Machines (QSVMs), Quantum Neural Networks (QNNs), Variational Quantum Classifiers (VQCs), Quantum k-Means, and Quantum Generative Adversarial Networks (QGANs). Each approach contributes distinct advantages: supervised models excel at detecting structured attacks, unsupervised models reveal hidden anomalies in large datasets, and generative models simulate sophisticated threat scenarios for adversarial defence. This section highlights the major ML techniques adapted for QML and their applications in cybersecurity.
Machine Learning serves as the foundation for both classical and quantum approaches in cybersecurity. ML enables learning from data, detecting hidden patterns, and adapting to evolving threats. Within QML, these methods are done through quantum mechanics, allowing models to operate in higher-dimensional feature spaces. It allows models to optimise more efficiently, and, in some settings, achieve greater accuracy in classification or clustering tasks. Core ML, including supervised learning, unsupervised learning, and generative modelling, when applied to the quantum domain, result in algorithms such as QSVMs, QNNs, VQCs, Quantum k-Means, and QGANs.  Structured attacks are detected well by supervised models, and hidden anomalies in large datasets are detected by unsupervised learning. Generative models can simulate complex and dynamic threat scenarios for adversarial defence. This section highlights such major ML techniques adapted for QML and their applications in cybersecurity.

\subsection{Supervised Learning}
% In cybersecurity, supervised learning \cite{t1} plays a key role in detecting and classifying threats. Supervised learning involves training models on labeled attack data. Quantum computing enhances this by enabling faster and more effective analysis of high-dimensional and complex threat patterns. While classical algorithms such as neural networks and SVMs perform well, they struggle with scalability. They also struggle when trained on very large datasets. Quantum-enhanced supervised models address these issues, making threat detection more efficient and accurate.
% In cybersecurity, supervised learning \cite{t1} plays a key role in detecting and classifying threats. Supervised learning involves training models on labeled attack data. Quantum computing enhances this by enabling faster and more effective analysis of high-dimensional and complex threat patterns. While classical algorithms such as neural networks and SVMs perform well, they struggle with scalability. They also struggle when trained on very large datasets. Quantum-enhanced supervised models address these issues, making threat detection more efficient and accurate.
Supervised learning has become a key technique for cyber defence, where models trained on labelled attack and benign traffic samples automatically detect and classify diverse threat types. Classical supervised algorithms like Neural Networks and SVMs can achieve strong performance, but their training and inference costs grow rapidly with data volume and feature dimensionality, making scalability difficult. Quantum computing can process high-dimensional, complex threat signatures more efficiently, and explore richer decision boundaries and patterns that may be difficult for classical methods to capture at scale. By combining supervised learning with quantum resources, these models aim to deliver more efficient and accurate threat detection pipelines. This is particularly true in scenarios involving massive, rapidly changing cyber telemetry where classical approaches begin to struggle.

\subsubsection{Quantum Support Vector Machines}
% Distinguishing malicious activity from normal behavior is often challenging because the differences can be subtle. Quantum Kernel Support Vector Machines (QSVMs) approach this by mapping data into quantum feature spaces, which makes patterns easier to identify while also accelerating computation \cite{t2}.  
% In \cite{y1}, a QSVM-based framework was introduced to detect False Data Injection Attacks (FDIAs) in Power Distribution Systems (PDS). Quantum feature mapping enabled the model to efficiently spot small anomalies within high-dimensional data, leading to an accuracy of 94.67\%. It also achieved an AUC of 0.98 for ramp attacks and 0.97 for random attacks on the ROC curve, surpassing classical SVMs in both accuracy and speed.  
QSVMs address a key problem in distinguishing malicious activity from normal behaviour. The differences in these activities can be very subtle. QSVMs map data into quantum feature spaces, making patterns easier to identify, while also accelerating computation for enhanced performance \cite{t2}.  
Janak et al.\cite{y1} proposed a QSVM-based framework to detect False Data Injection Attacks (FDIAs) in power distribution systems. The model was able to surpass classical SVMs in both accuracy and speed. It could efficiently spot small anomalies within high-dimensional data by quantum feature mapping, which led to an accuracy of 94.67\%. It also achieved an AUC of 0.98 for ramp attacks and 0.97 for random attacks on the ROC curve.
\begin{figure*}[h]
\centering
\includegraphics[width=\textwidth]{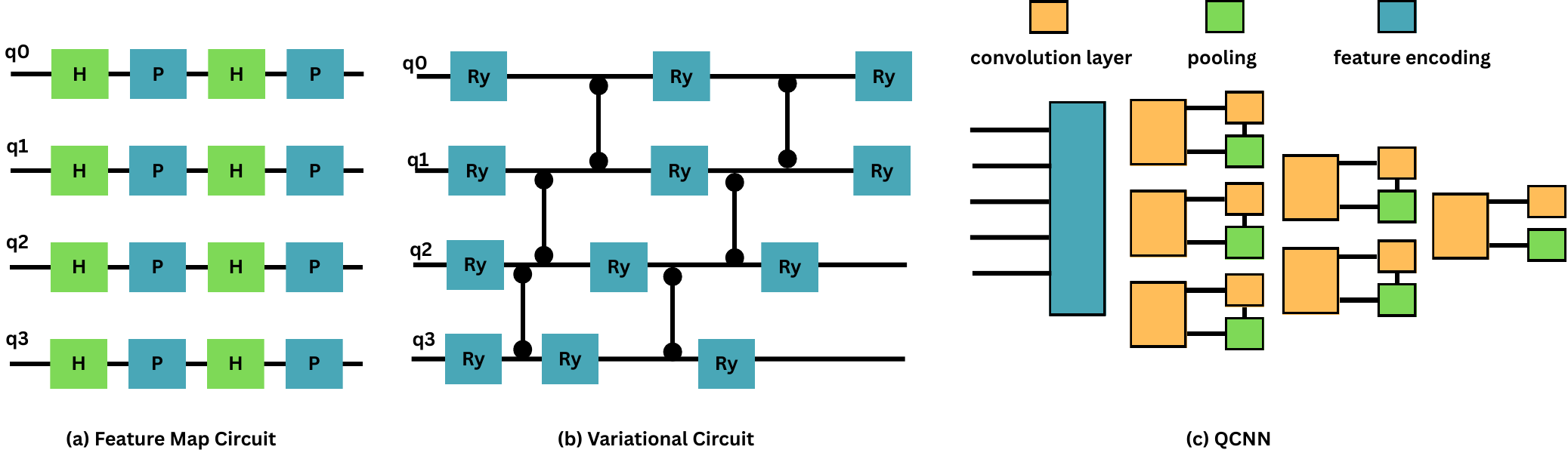}
\caption{Quantum circuits used in QML: (a) feature-map circuit, (b) variational circuit, and (c) Quantum Convolutional Neural Network (QCNN) showing convolution, pooling, and feature encoding.}
\label{fig:qcnn}
\end{figure*}
% In another study \cite{y2}, a hybrid classical-quantum QSVM was used to counter gradient-free adversarial attacks. By leveraging noise-robust quantum kernels, the model reduced adversarial susceptibility by 40–60\% on cybersecurity datasets while maintaining strong performance on clean data. This demonstrated QSVM’s ability to provide adversarial resilience in practical scenarios.
Kejriwal et al.\cite{y2} demonstrated the ability of QSVMs to provide resilience to adversaries in practical deployments. A hybrid classical-quantum QSVM was used to counter gradient-free adversarial attacks.  By leveraging noise-robust quantum kernels, adversarial susceptibility got reduced by 40–60\% on cybersecurity datasets while maintaining strong performance on clean data.

\subsubsection{Quantum Neural Networks}

% Complex attack vectors often require models that generalize across diverse and dynamic datasets. Quantum Neural Networks (QNNs), including Quantum Convolutional Neural Networks (QCNNs), provide advanced pattern recognition and have shown effectiveness for intrusion detection and malware analysis.
QNNs and QCNNs provide advanced pattern recognition. They have been shown to be especially effective for intrusion detection and malware analysis, which require models that generalise across diverse and dynamic datasets.

% In \cite{y3}, researchers developed an intrusion detection system using QNNs to classify DDoS attacks. The model leveraged quantum parallelism and superposition to process data from the CIC\mbox{-}DDoS 2019 dataset and achieved an accuracy of 92.63\%. Similarly, \cite{y4} proposed a QNN framework for real-time malware detection. The system comprised three stages: quantum-based feature extraction (QFT and QFM), classification with a VQC-driven QNN, and integration with a real-time analysis pipeline. Using the Malware DB dataset, the model achieved 95\% accuracy, outperforming classical approaches and other quantum baselines, while maintaining low latency suitable for real-time operation.
K{\"u}{\c{c}et al.\cite{y3} built an intrusion detection system using QNNs to classify DDoS attacks. They trained and tested their framework on the CIC-DDoS 2019 dataset. Their architecture utilised quantum parallelism and superposition for faster pattern processing and achieved an accuracy of 92.63\%.
Bikku et al.\cite{y4} designed a QNN-based framework for real-time malware detection. Their architecture had three main stages: quantum feature extraction, classification with a variational quantum circuit, and connection to a real-time analysis pipeline. They evaluated their architecture on the Malware DB dataset and achieved a 95\% accuracy, outperforming classical models and other quantum baselines. They achieved all this while maintaining low latency. Thus, they demonstrated that their architecture is suitable for real-time deployment.

\subsubsection{Variational Quantum Classifiers}
% VQCs adapt to evolving attack strategies by combining parameterized quantum circuits with classical optimizers in hybrid feedback loops. The work in \cite{r18} extended classical variational algorithms into quantum workflows by encoding classical data into quantum states, processing them with variational circuits, and optimizing parameters for improved classification. Paper \cite{r7} compared several hybrid quantum\mbox{-}classical binary classifiers on a DGA botnet dataset, with a focus on VQCs. The results highlighted both their potential and current limitations, and also introduced the Quantum Hoeffding Tree Classifier (QHTC) as an alternative.
VQCs combine parameterised quantum circuits with classical optimisers in hybrid feedback loops. Thus, they can adapt to evolving attack strategies. Abreu et al.\cite{r18}  integrated classical variational algorithms into quantum workflows by encoding classical data into quantum states, processing them with variational circuits, and optimising parameters for improved classification. Tehrani\cite{r7} compared several hybrid quantum-classical binary classifiers on a DGA botnet dataset, with a focus on VQCs. They also pitched QHTC as an alternative.

Abdur Rahman et al.\cite{y5} performed cyberattack detection using VQCs on the NSL-KDD dataset. They reduced the features from 8 to 3 using PCA. Their EfficientSU2 circuit with the COBYLA optimiser achieved 93\% training and 90\% test accuracy. They were able to beat both classical SVMs and other VQC implementations. Thus, they confirmed that fine-tuned VQCs can deliver strong cybersecurity performance. Figure \ref{fig:qcnn} illustrates the circuits underlying these approaches.

\subsection{Unsupervised Learning (Clustering)}
% Cybersecurity logs and network traffic often consist of vast amounts of unlabeled data. This dta may have hidden anomalies. Unsupervised learning methods help uncover these structures. While classical techniques like k-means and PCA are widely used, they struggle with scalability and complex high-dimensional data. Quantum computing, through parallelism and entanglement, addresses these limitations by making clustering faster and more insightful.
Quantum Computing makes clustering faster and more insightful using parallelism and entanglement. Thus, they overcome the limitations of classical unsupervised learning techniques like k-means and PCA, such as scalability and complex high-dimensional data. This is important because cybersecurity logs and network traffic often consist of vast amounts of unlabeled data.

\subsubsection{Quantum k-Means Clustering}
% Quantum k-means leverages quantum computing to speed up distance calculations and improve cluster mapping.  
% In \cite{y6}, researchers proposed a QML framework for clustering cybersecurity vulnerabilities from the 2022 CISA Known Exploited Vulnerabilities catalog. Using QCSWAPk-means and QkernelK-means, the study achieved better clustering metrics (Silhouette score of 0.491 and Davies-Bouldin index below 0.745) than classical k-means, producing more precise vulnerability groupings across severity and product categories.  
% Another approach, QALO-K \cite{y7}, combined classical k-means with a quantum-inspired ant lion optimization algorithm to overcome k-means’ limitations with initialization and local optima. By using quantum encoding and revolving gates, QALO-K achieved convergence improvements and reached over 98\% detection accuracy across multiple datasets, including intrusion detection scenarios.
Quantum k-means uses quantum computation to accelerate distance calculations, and thus they aim to produce more accurate and efficient cluster assignments. Maouaki et al.\cite{y6} proposed a framework for QML clustering for cybersecurity vulnerabilities. The dataset they targeted was the 2022 CISA Known Exploited Vulnerabilities(K-EV) Catalogue. They used the QCSWAP k-means and QKernel k-means methods and outperformed classical k-means with a Silhouette score of 0.491 and a Davies-Bouldin index below 0.745. They generated more meaningful clusters based on severity levels and affected products.

Similarly, Chenet al.\cite{y7} proposed a QALO-K hybrid clustering method combining classical k-means with a quantum-inspired ant lion optimisation strategy. They used quantum encoding and revolving gates for enhanced search and convergence and addressed k-means issues such as poor initialization and local minima traps. They thus achieved more than 98\% detection accuracy on multiple datasets, including intrusion detection datasets.

\subsection{Generative Adversarial Networks}
% GANs are commonly used to generate synthetic attack data for training robust detection models. Their quantum counterparts, QGANs, exploit quantum circuits to better capture complex distributions.  
% In \cite{y8}, a QGAN was developed with IBM’s Qiskit to support intrusion detection by generating atypical traffic patterns. With a complexity of $O(poly(n))$, the system enhanced security defences by identifying malicious packets more effectively than classical methods.  
% Similarly, \cite{y9} proposed a hybrid QGAN framework with a quantum generator and a classical discriminator, evaluated on the NSL\mbox{-}KDD dataset. After PCA-based preprocessing, the model achieved an accuracy of 0.937 and an F1\mbox{-}score of 0.9384. Performance improved with additional generator repetitions and larger discriminators, indicating the framework’s ability to balance accuracy and realism while maintaining stable training.
GANs can be used to generate synthetic attack data for training robust detection models. Quantum GANs use quantum circuits to capture complex distributions better than their classical counterparts. Rahmanet al.\cite{y8} developed a QGAN using Qiskit for intrusion detection. They achieved a complexity of $O(n)$ and the system enhanced security defenses by identifying malicious packets more effectively than classical methods for generating extraordinary traffic patterns. Cirilloet al.\cite{y9} also proposed a hybrid QGAN framework. They only utilised quantum technology in the generator and kept the discriminator block based on classical methods. They evaluated their framework in the NSL-KDD dataset.  The model achieved an accuracy of 0.937 and an F1 score of 0.9384. As they repeated the generator blocks and increased the discriminator size, they found improved performance. They indicated that their framework could balance accuracy and realism while maintaining stable training.

\section{QML for Cloud Computing Security}
% Cloud computing has transformed data storage, processing, and access, but     it has also introduced complex and evolving security challenges. With the prospect of quantum-enabled attacks, traditional cybersecurity approaches are increasingly inadequate, motivating stronger quantum-secure solutions. Research in this area examines how QML and related quantum technologies can be integrated with cloud platforms to deliver proactive, resilient, and scalable security frameworks. Collectively, these studies indicate that quantum-enhanced models can address limitations of classical methods to safeguard sensitive information and critical infrastructure.
Cloud computing centralises massive data and services on shared, virtualised infrastructure, creating risks like data breaches, insecure APIs, misconfigurations, and insider threats. This introduces complex and dynamic security challenges. As adversaries begin to explore quantum-capable attacks, many conventional cryptographic and defensive mechanisms are expected to become fragile or obsolete. Thus, there is a need to move beyond purely classical cybersecurity strategies. This threat landscape has inspired interest in defences that are quantum-enhanced and can withstand both current and future adversarial tactics, including those equipped with scalable quantum hardware. Recent works have investigated how quantum machine learning and other quantum technologies can be embedded into cloud-based infrastructures to enable more proactive, adaptive, and scalable security services, rather than relying on static or ad hoc protections. These studies suggest that quantum-enhanced security models can overcome several key limitations of classical techniques, offering stronger protection for sensitive data.

Zeng et al.\cite{l4} proposed a secure framework for Quantum Federated Learning using third-party quantum servers that allow classical users to participate. They used Quantum Homomorphic Encryption as the core mechanism, enabling computation on encrypted quantum data, therefore preserving the contents, not revealing them. The framework introduced the Quantum Circuit Random Reconstruction Homomorphic Encryption Algorithm (QCRRA). This algorithm encrypts variational circuits locally before sending them to the remote quantum server. This way, they protect both the user data and the structure of the model itself. They evaluated on a binary classification task using ad-hoc, MNIST and CIFAR-10 datasets. Their encrypted model, referred to as Encrypted Variational Quantum Circuit ($En\_VQC$), achieved an accuracy comparable to that of unencrypted VQCs. In several cases, $En\_VQC$ converges faster and reduces communication overhead. Therefore, QCRRA provides a theoretically infinite ciphertext space, providing much better strength and protection against ciphertext-based attacks. Also, their design maintained usability and model performance while ensuring data and model privacy in federal quantum learning.

% Paper \cite{l5} presented an Ensemble Intrusion Detection Model for Cloud Computing Using Deep Learning (EICDL) to counter emerging quantum-based attacks by enhancing intrusion detection. The architecture integrated deep learning with classical machine-learning models to improve accuracy and reliability. Evaluated on KDDcup 1999, UNSW\mbox{-}NB15, and NSL\mbox{-}KDD, the system achieved a recall of 92.14\%, surpassing existing intrusion detection methods. Reported results included improvements in detection precision, efficiency, and the identification of follow-up attacks relative to standalone models.
Salvakkam et al.\cite{l5} introduced EICDL, which stands for Ensemble Intrusion Detection Model for Cloud Computing using Deep Learning. They designed it to strengthen defences against machine quantum-enabled attacks. Their architecture combines deep learning models with classical machine learning classifiers, thereby increasing both accuracy and reliability compared to stand-alone methods. Their framework was able to achieve improvements in detection precision, efficiency, and ability to identify follow-up attacks. They evaluated it on KDD-CUP-1999, UNSW-NB15, and NSL-KDD datasets. The architecture achieved a recall of 92.14\%, outperforming existing intrusion detection baseline models. This analysis showed promise that hybrid deep learning and classical ML architectures can better secure cloud environments.

Gupta et al.\cite{L6} presented a quantum machine learning–based malicious user prediction (QM-MUP) scheme for identifying attackers in cloud environments before they gain access to sensitive data. They use QNNs trained on simulated user behaviour patterns, using qubits and quantum gate operations to capture richer, high-dimensional relationships in access and activity profiles than classical models typically can. QM-MUP achieved a 33.28\% improvement in overall system security relative to existing prediction approaches. They also achieved consistently high detection accuracy and were able to reliably flag malicious users. Complementing this predictive line of work, Kaliyamorthy et al.\cite{h1} proposed QMLFD-RSA, a cryptosystem designed to strengthen data integrity and security for public cloud storage scenarios. Their scheme combines the Quasi-Modified Lévy Flight Distribution (QMLFD) algorithm with the RSA public-key framework. QMLFD was used to improve key or parameter generation, while the RSA role ensured compatibility with existing infrastructures. The authors reported a packet delivery rate of 97.3\%, a throughput of 800 kbps, and a Time Complexity value of 97.3\%. Compared with baseline schemes such as RSA and DES, their results indicate that QMLFD-RSA offers significantly reduced time complexity. It also shows strong delivery and throughput characteristics, yielding a more efficient and secure option for cloud data integrity and communication.

Hybrid quantum–classical architectures built on QML have sparked particular interest for cloud management. Researchers claim that these hybrids lower runtime, energy use, and scaling stress compared to classical AI tools in hosted environments. Maddali\cite{h2} proposed a  framework that tries to push this promise into overcoming latency, inefficiency, and scalability bottlenecks in real-world cloud operations. The model combines quantum kernel methods, quantum Boltzmann machines, and variational quantum circuits within a broader classical framework for governance. The primary focus of their study was on anomaly detection and data consistency checks over cloud-hosted datasets flowing through shared clouds. The authors evaluated their framework on datasets drawn from finance, healthcare, and IoT sectors, and the results indicated up to a 10x speedup in anomaly detection and data quality verification relative to classical AI baselines. Execution times reduced by as much as 80\% and performance-conversion overheads cut by roughly 70\%, thus supporting the claims of improved speed, energy efficiency, and scalability. Despite these gains, Maddali and other studies remarked that hardware immaturity, deployment costs, and interoperability with existing cloud stacks remain obstacles for large-scale adoption of QML-based governance solutions. 
Broadly, the literature on QML in cloud security spans secure federated learning, quantum-safe and homomorphically encrypted analytics, proactive malicious-user prediction schemes, and quantum-enhanced data integrity and quality frameworks. Across these domains, QML-based approaches frequently outperform classical models on key metrics such as detection accuracy, computational efficiency, and scalability under realistic cloud workloads. These findings suggest that quantum-enhanced cloud security is gradually shifting from largely theoretical proposals to practically viable architectures, laying the groundwork for more secure, intelligent, and resilient cloud infrastructures.

\subsection{Future potential in cloud computing}
In cloud settings, QML must move from demonstrations to real services by supporting secure circuit submission usage quotas and auditable shot budgets. It must also create quantum-native patterns for multi-tenant use with strict isolation. Additionally, QML must provide serverless interfaces for variational inference and kernel execution, with clear latency and support for cross-targets. As far as privacy-preserving learning is concerned, it should combine quantum-predicted learning with post-quantum cryptography and local differential privacy, along with confidential computing. 

QML for the cloud should have explicit tracking of privacy versus utility data for encrypted data and model updates in regulated sectors. Reliability engineering should include fallback paths to classical circuit detectors, disaster recovery for analytics pipelines, also it should have circuit-level retries and cross-region circuit application. It should also have chaos-testing focused on quantum job queues and contributions. Resource and cost management should have cost models that link spending to incidents prevented, detection made, and service-level goals, with control of batch sizes, repartition counts, and error-mitigation processes. It should also have schedulers that decide optimal placement across simulators and hardware. As far as security is concerned, attestation for encrypted in-use analytics is a must. Also, audit evidence aligned to compliance frameworks should be provided. Integrating IAM, Key Management, SIEM, SOAR tools, and Zero Trust policies would also help in security and compliance requirements. 

As far as the edge cloud partnership strategy is concerned, feature extraction and privacy filters must be done at the edge, quantum kernels or variational layers in the cloud. We should measure end-to-end latency and reliability in distributed departments. Feature operability is also must, where standard APIs for data encoding, circuit specs, and telemetry are needed. Also needed is a benchmark source for contemplated security workloads that reports accuracy, latency, throughput, energy, cost, and robustness. A summary of future directions of research discussed above have been provided in Figure \ref{fig:future}.

\section{Limitations}
\label{sec6}
While QML is promising for cybersecurity, its deployment in the real world is hindered by several practical and fundamental obstacles. As per current works, the technology is still in a formative stage.  This is true for both the quantum hardware as well as the algorithmic frameworks around it, hindering its scalability, dependability and adoption.  The testing in operational environments has been very limited. Noisy Intermediate-Scale Quantum (NISQ) hardware
restrictions and training inefficiencies introduce further complications. Some additional concern,s such as lack
of interpretability, high costs, unclear regulatory standards, and
the dual-use nature of quantum technologies, have also been raised in the course of shifting from research prototypes to
full-scale deployments. These limitations have been discussed in detail in the subsections that follow.

\subsection{Hardware constraints \& NISQ limitations}

% QML is promising for intrusion detection, anomaly recognition, and malware analysis, yet reliance on NISQ devices creates substantive barriers. Present\mbox{-}day machines exhibit limited qubit counts, short coherence times, and high noise susceptibility. These factors reduce circuit stability, degrade model precision, and undermine threat\mbox{-}detection reliability. In security tasks, noisy outputs can cause false positives or missed detections in real time, eroding trust. A Systematic Mapping Study (SMS) by \cite{t3} reviewed these challenges for supervised QML, covering hardware options, datasets, optimization strategies, and generalization behavior, with benchmarks including MNIST and IRIS.
While  QML shows promise in intrusion detection, anomaly recognition, and malware analysis, it is limited by the current state of NISQ devices. Current machines have limited qubits, are highly susceptible to noise, and have short coherence times. All these factors lead to low circuit stability, low model precision, and undermined threat detection reliability. In the domain of security, the false positives produced by these noisy outputs harm trust. In their Systematic Mapping Study (SMS), Khanalet al.\cite{t3} analysed these challenges for supervised quantum machine learning (QML), evaluated on widely used datasets such as MNIST and IRIS. They focused on the choice of hardware, datasets, optimisation strategies, and generalisation.

\subsection{Optimization challenges \& barren plateaus}
% QML models used in cybersecurity, including QNNs and VQCs, face optimization issues that affect reliability for detection and response. Barren plateaus, vanishing gradients, and unstable convergence limit scalability and consistency \cite{t4}. Studies such as \cite{r8} and \cite{r11} investigate mitigations including data re\mbox{-}uploading and sparse parameterization. These techniques help in practice but introduce additional computation, complexity, and reduced interpretability, complicating deployment where timely and stable convergence is crucial.
QML models used in cybersecurity, such as QNNs and VQCs, suffer optimisation issues. These include barren plateaus, vanishing gradients, and unstable convergence. This limits their scalability and consistency\cite{t4}. Suryotrisongkoet al.\cite{r8} and Barru{\'e}et al.\cite{r11} discuss possible solutions such as data re-uploading and sparse parameterisation, but these solutions introduce additional computation, complexity, and reduced interpretability, complicating their deployment.

\subsection{Limited real\mbox{-}world testing}
% Evidence for QML in cybersecurity is still largely based on controlled evaluations. Despite promising reports, for example a 91.2\% detection rate for DGA botnets in \cite{r9}, most studies use static, pre\mbox{-}processed datasets and simulated settings rather than live network environments. Without field validation, concerns remain about robustness under noisy and adversarial conditions. Models trained on curated data may fail against adaptive attacks, evolving malware, or high\mbox{-}volume dynamic traffic.
The testing for QML in cybersecurity has largely been based on controlled evaluations. There have been some promising reports, such as the 91.2\% detection rate for DGA botnets in \cite{r9}. However, the majority of existing works have used static, pre-processed datasets and simulated settings rather than live network environments. They assume ideal conditions, which is not the case in real-world deployments that are infested with noise and adversarial conditions. Models trained on idealistic, curated data may fail when faced with adaptive attacks, evolving malware, or high-volume, dynamic traffic.

\subsection{Explainability \& transparency gaps}
% QML systems for security are typically opaque, offering limited interpretability. This is problematic in operational contexts where analysts must justify why traffic or activity is labeled malicious. Early methods such as Permutation Importance (PI) and Accumulated Local Effects (ALE) have been explored \cite{r19}, but they can be resource intensive and inconsistent across implementations. The absence of reliable explainability reduces analyst trust and hampers incident handling, slowing adoption and regulatory approval in high\mbox{-}stakes environments.
QML systems for cybersecurity are mostly black-boxes. They offer limited interpretability. This becomes problematic in operations where the justification of the labelling of an activity as malicious is demanded. Some approaches, such as Permutation Importance and Accumulated Local Effects, have been explored in the past, but they have been shown to be resource-intensive and inconsistent across implementations. The lack of explainability hinders the adoption of QML due to a lack of analyst trust.  

\subsection{Quantum vs. classical crypto risk}
% Advances in quantum technology both strengthen and threaten defences. Large\mbox{-}scale quantum computers endanger classical cryptographic schemes while also enabling stronger mechanisms. As emphasized in \cite{r6}, migration toward post\mbox{-}quantum cryptography is urgent for infrastructures that rely on classical encryption. Introducing QML without synchronized cryptographic migration could expose sensitive data to future quantum\mbox{-}enabled attacks. QML adoption should therefore be coordinated with broader resilience strategies.
As with any technology, quantum computations can both be used and misused. QML is powerful not in defence, but has powerful capabilities in offence. Large-scale quantum computers endanger classical cryptographic schemes and provide hard-to-counter mechanisms. Marengoet al.\cite{r6} show that migration toward post-quantum cryptography is urgent for infrastructures that rely on classical encryption. If cryptography is not simultaneously migrated from classical to quantum architectures,  it can expose sensitive data to future quantum-based attacks. Thus, QML adoption must be accompanied with broader resilience strategies. 

\subsection{High cost \& access barriers}
% Practical deployment faces high costs and limited access. Building QML pipelines requires specialized SDKs, controlled access to hardware via providers such as IBM Q and AWS Braket, and multidisciplinary expertise spanning AI, cybersecurity, and quantum engineering. Studies including \cite{r7} and \cite{r14} note that these requirements favor well\mbox{-}funded organizations, limiting participation by smaller enterprises and under\mbox{-}resourced regions and potentially widening the cybersecurity divide.

To implement QML pipelines at operational scale, teams must manage high costs, limited quantum hardware availability, and complex toolchains. Studies like \cite{r7} and \cite{r14} show that such demands benefit only well-funded organizations, leaving smaller ones and low-resource regions at a disadvantage, restricting broader participation and creating a divide in cybersecurity.

\subsection{Lack of regulatory frameworks}
% The absence of consistent regulatory frameworks complicates QML adoption. Issues identified in \cite{r24} include accountability for algorithmic outcomes, cross\mbox{-}border data handling, privacy protection, and alignment with ethical AI principles. In cybersecurity, these gaps risk fragmented standards and uneven protections. Without coordinated oversight, premature integration could amplify risks of bias, misuse, and inconsistent security enforcement.
% There is a lack of regulatory frameworks in cybersecurity that hinder QML adoption. \cite{r24} identify issues such as accountability for algorithmic outcomes, cross-border data handling, privacy protection, and alignment with ethical AI principles. These gaps risk incoherent standards and biased protections. If done without coordinated oversight, premature integration could increase the risks of bias, misuse, and inconsistent security enforcement.
Current cybersecurity regulations are insufficient to support responsible QML deployment, and limit its safe adoption. As noted in \cite{r24}, challenges persist in areas such as accountability for algorithmic decisions, international data governance, privacy protection, and ethical AI compliance. These shortcomings may result in inconsistent standards and biased protections. Also, early adoption without strong oversight could heighten risks of misuse and uneven security enforcement.

\section{Future Work}
As QML for cybersecurity continues to become increasingly explored, the limitations outlined above highlight the need for more operationally grounded research practices that rigorous and transparent. Addressing these gaps will not only improve scientific reliability but also strengthen real-world applicability across adversarial environments and reproducibility. The following directions outline practical steps that can be taken to build a more robust and secure foundation for future work.\vspace{1em}

\begin{figure*}[!t]
\centering
\includegraphics[trim=0 4cm 0 0, clip, width=\textwidth]{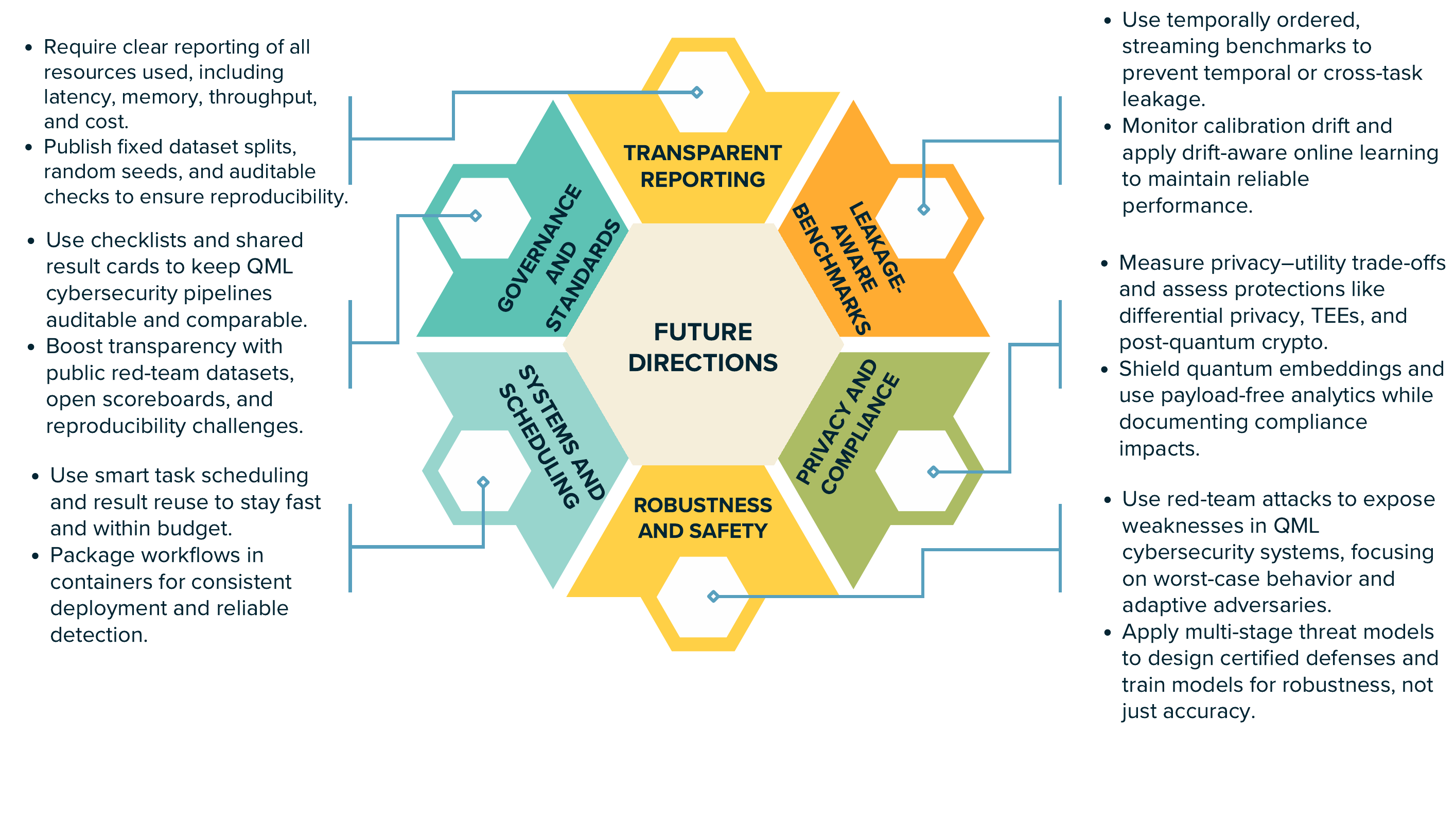}
\caption{Summary of future research directions}
\label{fig:future}
\end{figure*}

\noindent\textbf{Transparent reporting and leakage-aware benchmarks:}
% Time-ordered, streaming benchmarks can be developed to mirror operational telemetry and SOC workflows. Each study can include complete resource accounts—query counts, shot budgets, latency, throughput, memory footprint, and monetary cost—alongside accuracy, calibration quality, confidence intervals, and drift statistics. Reproducible splits and seeds, together with auditable cross-task leakage checks, enable faithful, end-to-end replication.
There is a need for benchmark suites that capture temporally ordered and streaming data, accurately representing the telemetry and workflow behaviour of real Security Operations Centres. In QML-based cybersecurity pipelines, such benchmarks are particularly important because temporal leakage can lead to unrealistic advantages when evaluating hybrid quantum–classical classifiers. Additionally, for every experiment, full resource usage must be reported. Resources like the number of queries issued, short budgets, latency, throughput, memory footprint, and total monetary cost need to be reported. Performance metrics such as accuracy and calibration quality, confidence intervals, and drift statistics over time also have to be studied. 
By publishing fixed dataset splits and random seeds, reproducibility can be ensured. It can also be ensured by producing auditable checks preventing cross-task or temporal leakage. Thus limited qubit lifetimes, gaps in the interpretability of models, optimisation bottlenecks, and limited hardware remain problems, but the pace of current research indicates progress toward practical and high-fidelity deployment of QML in cybersecurity, constrained operators). Pre-mitigation and on-the-fly shot scheduling can be compared under equal budgets, while longitudinal tracking of calibration drift quantifies accuracy–latency–cost trade-offs across versions. Drift-aware online learning with warm starts and budgeted hyperparameter optimization can be implemented to maintain performance under shift.
Specific properties related to the hardware and data, such as gate sets, noise levels, restricted operations, etc., can be kept in consideration while choosing encodings and shallow-depth prompting. In QML-driven intrusion detection, these encoding choices may also influence how robust the model is to adversarial manipulation. The drift of calibrations in hardware should be continuously monitored when quantifying accuracy-delay-cost comparisons. To maintain performance under shift, drift-aware online learning with warm starts and budgeted parameter tuning can be implemented. On-the-fly shot scheduling, and doing so before correction, can be compared under equal budgets.

\noindent\textbf{Governance, tooling, and community standards:}
% Checklists covering dataset provenance, prompt/version pinning, and leak-prevention reviews can improve scientific hygiene. Standardized result cards and failure taxonomies enable apples-to-apples comparisons, while open red-team corpora and public scoreboards foster transparency. Community reproducibility challenges and recognition badges can further incentivize robust, end-to-end evaluations.
Structured checklists that record where datasets came from, which prompt or model version was used, and whether any training leaks were prevented can make research cleaner and more trustworthy. In the context of QML for cybersecurity, such governance helps ensure that hybrid quantum pipelines remain auditable and reproducible. When results are reported through a shared result card format and failures are labeled using the same taxonomy, systems can be compared fairly instead of loosely. Public red-team datasets and open scoreboards add transparency by showing how models actually behave under stress tests, rather than only in ideal settings. Finally, community challenges and recognition badges can encourage researchers to fully reproduce and validate entire pipelines instead of stopping at headline numbers, promoting healthy research.

\noindent\textbf{Privacy-preserving analytics and compliance:}
% Encrypted-traffic analytics and telemetry aggregation without payload visibility can be advanced, complemented by evaluations of differential privacy, trusted execution environments, and post-quantum cryptography for data-in-use protections. Privacy–utility trade-offs can be measured across representative workloads, and domain-specific compliance implications can be documented to support audits and policy alignment.
Trade-offs in privacy and utility can first be measured across real workloads to understand how much functionality is lost when stronger protections are applied. These protections may include differential privacy, trusted execution environments (TEEs), and post-quantum cryptography. In QML-enabled threat detection systems, privacy-preserving computation is particularly relevant because quantum embeddings may expose structure in sensitive telemetry if not properly shielded. Each of these protections should be evaluated for how well they secure data while it is actively being used. Encrypted-traffic analytics and metadata-level telemetry aggregation can then be advanced to operate without inspecting payloads at all. Finally, the compliance impact of these techniques can be documented for domain-specific audits and policy alignment.

\noindent\textbf{Robustness and safety evaluation:}
% Red-team frameworks can exercise evasion, poisoning, and cross-domain (including quantum-aware) attacks to stress-test the stack. Certified defences and robustness-oriented training objectives may be derived under formal, multi-stage threat models. Evaluation can prioritize failure analyses and worst-case behavior, with metrics designed for adaptive adversaries rather than averages alone.
Red-team setups can deliberately launch different kinds of attacks such as evasion, data poisoning, and even quantum-aware cross-domain attacks, to proactively break the system and reveal weak points. For QML models used in cybersecurity, such tests can evaluate whether quantum feature maps or variational circuits introduce new attack surfaces. Instead of relying on average performance, evaluation should focus on worst-case behavior, failure modes, and how the system reacts when the attacker adapts. Then, formal multi-stage threat models can be used to design certified defences and to set training objectives that explicitly target robustness rather than just accuracy.

\noindent\textbf{Systems, partitioning, and scheduling:}
% Partitioning/placement, batching, and caching can be optimized to satisfy latency-sensitive requirements without violating cost or energy constraints. Schedulers can be designed to respect mission-aligned budgets for shots and compute while meeting explicit service-level objectives for throughput, tail latency, and time-to-detect. Containerized reference pipelines integrated with continuous integration can validate data exchange and compilation artifacts for reliable deployment.
Performance can be improved by intelligently splitting tasks across machines, grouping them together, and reusing stored results, so that the systems remain fast and do not overspend energy and money unnecessarily. When QML components are part of a broader cybersecurity monitoring pipeline, this system-level tuning directly influences time-to-detect and alert reliability. While scheduling the jobs, the scheduler should follow a fixed budget (shots, compute time, etc.) but still hit targets like throughput, worst-case latency, and detection time. Containers can be used to package workflows for automated testing and to ensure that data formats and compiled code stay valid every time we deploy.

\section{Conclusions}

QML gives new hope for cybersecurity by combining quantum computation with machine learning. QML can improve detection accuracy, expose hidden anomalies, adapt to
evolving threats in real time, improving much over rule-based and traditional machine learning based systems. In this review, applications of QML across healthcare, malware and botnet detection, and telecommunication security have been surveyed. QSVMs, QNNs, VQCs, and QGANs based frameworks were shown to excel in both supervised and unsupervised settings. Although hybrid qunatum-classical models, in the current
NISQ era, have hardware constraints, strong potential still surfaced across studies. These findings highlight rapid progress of the field towards maturity. Limited qubit lifetimes, gaps in the interpretability of models, optimization bottlenecks, and limited hardware remain problems, but the pace of current research indicate progress toward practical and high-fidelity deployment of QML in cybersecurity. As experimental platforms and learning techniques continue to improve, QML is positioned to support more robust, scalable, and context-aware security systems capable of addressing increasingly complex threat landscapes.

\section*{Acknowledgments}
\thanks{This work was supported by CHANAKYA Fellowship Program of TIH Foundation for IoT \& IoE (TIH-IoT) received by Dr. Vinay Chamola under Project Grant File CFP/2022/027 and an ARC Discovery Project. }

\bibliographystyle{IEEEtran}
\bibliography{bibliography1.bib}

@article{r1,
  title={Quantum Machine Learning for Enhanced Cybersecurity: Proposing a Hypothetical Framework for Next-Generation Security Solutions},
  author={Hossain, Forhad and Hasan, Kamrul and Amin, Al and Mahmud, Shakik},
  year={2024},
  publisher={IADITI Editions}
}

@article{r2,
  author    = {Amit Awasthi},
  title     = {The Role of Quantum Machine Learning in Cybersecurity},
  journal   = {International Research Journal of Modernization in Engineering Technology and Science},
  year      = {2025},
  volume    = {7},
  number    = {1}
}

@article{r3,
  author  = {Juan, Alan and Ethan, Amelia},
  title   = {The Convergence of Quantum Computing and Deep Learning in Threat Modeling},
  journal = {Revista de Inteligencia Artificial en Medicina},
  volume  = {16},
  number  = {01},
  year    = {2025},
  url     = {https://redcrevistas.com/index.php/Revista}
}

@article{r4,
  author  = {Gupta, Kishu and Saxena, Deepika and Rani, Pooja and Kumar, Jitendra and Makkar, Aaisha and Singh, Ashutosh Kumar and Lee, Chung-Nan},
  title   = {An Intelligent Quantum Cyber-Security Framework for Healthcare Data Management},
  journal = {IEEE Transactions on Automation Science and Engineering},
  year    = {2024}
}

@incollection{r5,
  author    = {Sundaram, Aishwarya},
  title     = {Challenges and Opportunities in Quantum Computing in Healthcare},
  booktitle = {Quantum Computing for Healthcare Data},
  pages     = {91--118},
  year      = {2025}
}

@article{r6,
  author  = {Marengo, Agostino and Santamato, Vito},
  title   = {Quantum Algorithms and Complexity in Healthcare Applications: A Systematic Review with Machine Learning-Optimized Analysis},
  journal = {Frontiers in Computer Science},
  volume  = {7},
  pages   = {1584114},
  year    = {2025}
}

@phdthesis{r7,
  author = {Tehrani, Madjid Golparvaran},
  title  = {Quantum Cybersecurity Analytics: Evaluation of Hybrid QML for DGA Botnet Detection},
  school = {The George Washington University},
  year   = {2024}
}

@inproceedings{r8,
  author    = {Suryotrisongko, Hatma and Musashi, Yasuo},
  title     = {Evaluating Hybrid Quantum-Classical Deep Learning for Cybersecurity Botnet DGA Detection},
  booktitle = {Procedia Computer Science},
  volume    = {197},
  pages     = {223--229},
  year      = {2022}
}

@article{r9,
  author         = {Tehrani, Madjid and Sultanow, Eldar and Buchanan, William J. and Amir, Malik and Jeschke, Anja and Chow, Raymond and Lemoudden, Mouad},
  title          = {Enabling Quantum Cybersecurity Analytics in Botnet Detection: Stable Architecture and Speed-up through Tree Algorithms},
  journal        = {arXiv:2306.13727},
  year           = {2023}
}

@article{r10,
  author  = {Bikku, Thulasi and Chandolu, Suresh Babu and Praveen, S. Phani and Tirumalasetti, Narasimha Rao and Swathi, K. and Sirisha, U.},
  title   = {Enhancing Real-Time Malware Analysis with Quantum Neural Networks},
  journal = {Journal of Intelligent Systems \& Internet of Things},
  volume  = {12},
  number  = {1},
  year    = {2024}
}

@misc{r26,
  author        = {{Data Intelo}},
  title         = {Quantum AI Malware Detection Market Research Report 2024--2032},
  howpublished  = {\url{https://dataintelo.com/report/quantum-ai-malware-detection-market}},
  note          = {Accessed online},
  year          = {2024}
}

@inproceedings{r11,
  author    = {Barru{\'e}, Gr{\'e}goire and Quertier, Tony},
  title     = {Quantum Machine Learning for Malware Classification},
  booktitle = {Joint European Conference on Machine Learning and Knowledge Discovery in Databases},
  pages     = {245--260},
  publisher = {Springer Nature Switzerland},
  address   = {Cham},
  year      = {2023}
}

@incollection{r12,
  author    = {Kr{\'a}tk{\'a}, Eli{\v{s}}ka and G{\'a}bris, Aur{\'e}l G{\'a}bor},
  title     = {Quantum Computing Methods for Malware Detection},
  booktitle = {Machine Learning, Deep Learning and AI for Cybersecurity},
  pages     = {207--228},
  publisher = {Springer Nature Switzerland},
  address   = {Cham},
  year      = {2025}
}

@article{r13,
  author  = {Oyebode, Oyeyemi Abiola and Jimoh, Adam Akanmu},
  title   = {Quantum Cryptography in Telecommunication Systems: Securing Data Transmission Against Emerging Cyber Threats},
  journal = {International Journal of Computer Applications Technology and Research},
  month   = {February},
  year    = {2025}
}

@misc{r14,
  author = {Noah, Peter and Alexander, David},
  title  = {Quantum Machine Learning for Enhancing Cybersecurity in 5G Networks},
  year   = {2023},
  note   = {\url{https://www.researchgate.net/publication/390286351_Quantum_Machine_Learning_for_Enhancing_Cybersecurity_in_5G}}
}

@inproceedings{h3,
  author    = {Ergu, Yared Abera and Nguyen, Van-Linh and Lin, Po-Ching and Hwang, Ren-Hung},
  title     = {Q-Poison: Quantum Adversarial Attacks against QML-driven Interference Classification in O-RAN},
  booktitle = {2025 IEEE International Conference on Machine Learning for Communication and Networking (ICMLCN)},
  doi       = {10.1109/ICMLCN64995.2025.11140504},
  year      = {2025}
}

@inproceedings{t2,
  author    = {Bologna, Guido and Hayashi, Yoichi},
  title     = {QSVM: A Support Vector Machine for Rule Extraction},
  booktitle = {International Work-Conference on Artificial Neural Networks},
  pages     = {276--289},
  publisher = {Springer International Publishing},
  address   = {Cham},
  year      = {2015}
}

@article{t3,
  author  = {Khanal, Bikram and Rivas, Pablo and Sanjel, Arun and Sooksatra, Korn and Quevedo, Ernesto and Rodriguez, Alejandro},
  title   = {Generalization Error Bound for Quantum Machine Learning in the NISQ Era—A Survey},
  journal = {Quantum Machine Intelligence},
  volume  = {6},

  number  = {2},
  pages   = {90},
  year    = {2024}
}

@article{t4,
  author  = {Cunningham, Jack and Zhuang, Jun},
  title   = {Investigating and Mitigating Barren Plateaus in Variational Quantum Circuits: A Survey},
  journal = {Quantum Information Processing},
  volume  = {24},
  number  = {2},
  pages   = {48},
  year    = {2025}
}

@inproceedings{r18,
  author    = {Abreu, Diego and Rothenberg, Christian Esteve and Abel{\'e}m, Antonio},
  title     = {QML-IDS: Quantum Machine Learning Intrusion Detection System},
  booktitle = {2024 IEEE Symposium on Computers and Communications (ISCC)},
  pages     = {1--6},
  publisher = {IEEE},
  year      = {2024}
}

@article{l1,
  author    = {Karamchand, Gopalakrishna},
  title     = {Quantum Machine Learning for Threat Detection in High-Security Networks},
  journal   = {SAMRIDDHI: A Journal of Physical Sciences, Engineering and Technology},
  volume    = {17},
  number    = {2},
  pages     = {14--25},
  publisher = {The Author(s)},
  year      = {2025}
}

@article{v3,
  author  = {Kejriwal, Deepak Kumar and Goel, Anshul and Sharma, Ashwin},
  title   = {Advancing Adversarial Robustness in Cybersecurity: Gradient-Free Attacks and Quantum-Inspired Defenses for Machine Learning Models},
  journal = {International Journal of Innovative Science and Research Technology},
  volume  = {10},
  number  = {4},
  pages   = {54--65},
  year    = {2025}
}

@mastersthesis{y1,
  author = {Janak, Urmisha Reddy},
  title  = {Detection of Cyber Attacks on Power Distribution System Using {QSVM}},
  school = {Purdue University},
  year   = {2024},
  month  = {12},
  type   = {{M.Sc.} thesis},
  doi    = {10.25394/PGS.27976563.v1}
}

@article{y2,
  author  = {Kejriwal, Deepak Kumar and Goel, Anshul and Sharma, Ashwin},
  title   = {Advancing Adversarial Robustness in Cybersecurity: Gradient-Free Attacks and Quantum-Inspired Defenses for Machine Learning Models},
  journal = {International Journal of Innovative Science and Research Technology},
  volume  = {10},
  number  = {4},
  pages   = {54--65},
  year    = {2025}
}

@article{y3,
  author  = {K{\"u}{\c{c}}{\"u}kkara, Muhammed Yusuf and Atban, Furkan and Bay{\i}lm{\i}{\c{s}}, C{\"u}neyt},
  title   = {Quantum-Neural Network Model for Platform-Independent DDoS Attack Classification in Cyber Security},
  journal = {Advanced Quantum Technologies},
  volume  = {7},
  number  = {8},
  pages   = {2400084},
  publisher = {Wiley Online Library},
  year    = {2024}
}

@article{y4,
  author  = {Bikku, Thulasi and Chandolu, Suresh Babu and Praveen, S. Phani and Tirumalasetti, Narasimha Rao and Swathi, K. and Sirisha, U.},
  title   = {Enhancing Real-Time Malware Analysis with Quantum Neural Networks},
  journal = {Journal of Intelligent Systems and Internet of Things},
  volume  = {12},
  number  = {01},
  pages   = {57--69},
  year    = {2024}
}

@inproceedings{y5,
  author    = {Rahman, Md Abdur and Akter, Mst. Shapna and Miller, Emily and Timofti, Bogdan and Shahriar, Hossain and Masum, Mohammad and Wu, Fan},
  title     = {Fine-Tuned Variational Quantum Classifiers for Cyber Attacks Detection Based on Parameterized Quantum Circuits and Optimizers},
  booktitle = {2024 IEEE 48th Annual Computers, Software, and Applications Conference (COMPSAC)},
  pages     = {1067--1072},
  doi       = {10.1109/COMPSAC61105.2024.00144},
  year      = {2024}
}

@inproceedings{y6,
  author    = {Maouaki, Walid El and Innan, Nouhaila and Marchisio, Alberto and Said, Taoufik and Bennai, Mohamed and Shafique, Muhammad},
  title     = {Quantum Clustering for Cybersecurity},
  booktitle = {2024 IEEE International Conference on Quantum Computing and Engineering (QCE)},
  volume    = {02},
  pages     = {5--10},
  year      = {2024}
}

@article{y7,
  author  = {Chen, Junwen and Qi, Xuemei and Chen, Linfeng and Chen, Fulong and Cheng, Guihua},
  title   = {Quantum-Inspired Ant Lion Optimized Hybrid k-Means for Cluster Analysis and Intrusion Detection},
  journal = {Knowledge-Based Systems},
  volume  = {203},
  pages   = {106167},
 
  doi     = {10.1016/j.knosys.2020.106167},
 year    = {2020}
}

@inproceedings{y8,
  author    = {Rahman, Md Abdur and Shahriar, Hossain and Clincy, Victor and Hossain, Md Faruque and Rahman, Muhammad},
  title     = {A Quantum Generative Adversarial Network-based Intrusion Detection System},
  booktitle = {2023 IEEE 47th Annual Computers, Software, and Applications Conference (COMPSAC)},
  pages     = {1810--1815},
  year      = {2023}
}

@inproceedings{y9,
  author    = {Cirillo, Franco and Esposito, Christian},
  title     = {Intrusion Detection System Based on Quantum Generative Adversarial Network},
  booktitle = {Proceedings of the 17th International Conference on Agents and Artificial Intelligence (ICAART 2025)},
  pages     = {830--838},
  publisher = {SCITEPRESS--Science and Technology Publications, Lda.},
  year      = {2025}
}

@article{h1,
  author    = {Kaliyamoorthy, Priyadharshini and Ramalingam, Aroul Canessane},
  title     = {QMLFD Based RSA Cryptosystem for Enhancing Data Security in Public Cloud Storage System},
  journal   = {Wireless Personal Communications},
  volume    = {122},
  pages     = {755--782},
  doi       = {10.1007/s11277-021-08924-z},
  publisher = {Springer},
  year      = {2022}
}

@article{h2,
  author  = {Maddali, Raghavender},
  title   = {AI-Driven Quality Assurance in Cloud-Based Data Systems: Quantum Machine Learning for Accelerating Data Quality Metrics Calculation},
  journal = {International Journal of Scientific Research in Computer Science, Engineering and Information Technology},
  volume  = {8},
  number  = {4},
  pages   = {366--382},
  doi     = {10.32628/IJSRCSEIT},
  issn    = {2456-3307},
  year    = {2022}
}

@ARTICLE{l2,
  author={Al-Hawawreh, Muna and Hossain, M. Shamim},
  journal={IEEE Internet of Things Journal}, 
  title={A Human-Centered Quantum Machine Learning Framework for Attack Detection in IoT-Based Healthcare Industry 5.0}, 
  year={2025},
  volume={12},
  number={22},
  pages={46065-46074},
  doi={10.1109/JIOT.2025.3565687}}

@article{r21,
  title={Quantum algorithms for supervised and unsupervised machine learning},
  author={Lloyd, Seth and Mohseni, Masoud and Rebentrost, Patrick},
  journal={arXiv:1307.0411},
  year={2013}
}

@inproceedings{r24,
  author    = {Nokhwal, Sahil and Nokhwal, Suman and Pahune, Saurabh and Chaudhary, Ankit},
  title     = {Quantum Generative Adversarial Networks: Bridging Classical and Quantum Realms},
  booktitle = {Proceedings of the 2024 8th International Conference on Intelligent Systems, Metaheuristics \& Swarm Intelligence},
  pages     = {105--109},
  year      = {2024}
}

@article{l4,
  author  = {Zeng, Lin and Chang, Yan and Zhang, Xuejian and Xue, Weifeng and Zhang, Shibin and Yan, Lili and Gou, Zhijian},
  title   = {Distributed Machine Learning Based on Quantum Cloud with Quantum Homomorphic Encryption},
  journal = {Future Generation Computer Systems},
  volume  = {175},
  pages   = {108053},
  publisher = {Elsevier},
  year    = {2026}
}

@article{l5,
  author    = {Salvakkam, Dilli Babu and Saravanan, Vijayalakshmi and Jain, Praphula Kumar and Pamula, Rajendra},
  title     = {Enhanced Quantum-Secure Ensemble Intrusion Detection Techniques for Cloud Based on Deep Learning},
  journal   = {Cognitive Computation},
  volume    = {15},
  pages     = {1593--1612},
  doi       = {10.1007/s12559-023-10139-2},
  publisher = {Springer},
  year      = {2023}
}

@article{l6,
  author  = {Gupta, Rishabh and Saxena, Deepika and Gupta, Ishu and Makkar, Aaisha and Singh, Ashutosh Kumar},
  title   = {Quantum Machine Learning Driven Malicious User Prediction for Cloud Network Communications},
  journal = {IEEE Networking Letters},
  volume  = {4},
  number  = {4},
  pages   = {174--178},
  doi     = {10.1109/LNET.2022.3200724},
  year    = {2022}
}

@article{r25,
  author  = {Lloyd, Seth and Weedbrook, Christian},
  title   = {Quantum Generative Adversarial Learning},
  journal = {Physical Review Letters},
  volume  = {121},
  number  = {4},
  pages   = {040502},
  year    = {2018}
}

@inproceedings{Nguyen2024,
  author    = {Nguyen, Thien and Sipola, Tuomo and Hautam{\"a}ki, Jari},
  title     = {Machine Learning Applications of Quantum Computing: A Review},
  booktitle = {Proceedings of the 23rd European Conference on Cyber Warfare and Security (ECCWS)},
  publisher = {Academic Conferences International},
  doi       = {10.34190/eccws.23.1.2258},
  note      = {Funded by the European Union},
  year      = {2024}
}

@article{Eze2025,
  author    = {Eze, Lauren and Chaudhry, Umair B. and Jahankhani, Hamid},
  title     = {Quantum-Enhanced Machine Learning for Cybersecurity: Evaluating Malicious URL Detection},
  journal   = {Electronics},
  volume    = {14},
  number    = {9},
  pages     = {1827},
  publisher = {MDPI},
  doi       = {10.3390/electronics14091827},
  url       = {https://doi.org/10.3390/electronics14091827},
  year      = {2025}
}

@article{Ganapathy2025,
  author    = {Ganapathy, Venkatasubramanian},
  title     = {Quantum Machine Learning for Anomaly Detection in Cyber Security Audits},
  journal   = {Shodh Sari -- An International Multidisciplinary Journal},
  volume    = {4},
  number    = {1},
  pages     = {127--154},
  publisher = {International Council for Education Research and Training},
  doi       = {10.59231/SARI7784},
  year      = {2025}
}

@article{Kiranmai2025,
  author    = {Dornala, Sai Kiranmai and Senthilkumar, P.},
  title     = {A Comprehensive Survey on Intelligent Firewall-Based Malware Detection Using Proactive and Quantum Approaches},
  journal   = {International Journal of Computer Networks and Wireless Communications (IJCNWC)},
  volume    = {15},
  number    = {2},
  pages     = {1123--1133},
  publisher = {IRACST},
  issn      = {2250-3501},
  year      = {2025}
}

\vskip 10\baselineskip plus -1fil
% \vskip -1\baselineskip plus -1fil
% \vskip -1\baselineskip plus -1fil
% \vskip -1\baselineskip plus -1fil
% \vskip -1\baselineskip plus -1fil

\end{document}